\pdfoutput=1

\documentclass[11pt]{article}

\usepackage[preprint]{acl}

\usepackage{times}
\usepackage{latexsym}
\usepackage{amsmath} 
\usepackage{amssymb}
\usepackage{bm}

\usepackage[T1]{fontenc}

\usepackage[utf8]{inputenc}

\usepackage{microtype}

\usepackage{inconsolata}

\usepackage{graphicx}
\usepackage{booktabs}
\usepackage{multirow}
\usepackage{subcaption}

\usepackage[most]{tcolorbox}

\usepackage{listings}
\usepackage{setspace}
\usepackage{color}

\definecolor{isarblue}{HTML}{006699}
\definecolor{isarfaintblue}{rgb}{0.0, 0.75, 1.0}
\definecolor{isargreen}{HTML}{009966}
\definecolor{red}{HTML}{990000}
\definecolor{patriarch}{rgb}{0.5, 0.0, 0.5}

\lstdefinelanguage{isabelle}{%
    keywords=[1]{theory,type_synonym,datatype,fun,abbreviation,definition,proof,lemma,theorem,qed,corollary,have,hence,also,finally,ultimately,moreover,using,\{},
    keywordstyle=[1]\bfseries\color{isarblue},
    keywords=[2]{where,assumes,shows,fixes,and,begin,end,imports},
    keywordstyle=[2]\bfseries\color{isargreen},
    keywords=[3]{if,then,else,case,SOME,let,in,O},
    keywordstyle=[3]\color{isarblue},
    keywords=[4]{ATP},
    keywordstyle=[4]\it\color{patriarch},
    keywords=[5]{show,assume,obtain},
    keywordstyle=[5]\bfseries\color{isarfaintblue},
    keywords=[6]{<proof>},
    keywordstyle=[6]\color{yellow},
}

\lstdefinestyle{isabelle}{%
  language=isabelle,
  escapeinside={&}{&},
  columns=fixed,
  extendedchars,
  basewidth={0.5em,0.45em},
  basicstyle=\singlespacing\ttfamily,
  mathescape,
  morecomment=[s][\bfseries\color{red}]{(*}{*)},
  morecomment=[l][\bfseries]{####},
}

\definecolor{mybrown}{RGB}{128,64,0}

\title{Autoformalization in the Wild: Assessing LLMs on \\Real-World Mathematical Definitions}

\author{
  \textbf{Lan Zhang\textsuperscript{1}},
  \textbf{Marco Valentino\textsuperscript{2}},
  \textbf{Andr\'e Freitas\textsuperscript{1,3,4}}\\
  \textsuperscript{1}Department of Computer Science, University of Manchester, United Kingdom\\
  \textsuperscript{2}School of Computer Science, University of Sheffield, United Kingdom\\
  \textsuperscript{3}Idiap Research Institute, Switzerland\\
  \textsuperscript{4}National Biomarker Centre, CRUK Manchester Institute, United Kingdom\\
  \texttt{lan.zhang-6@postgrad.manchester.ac.uk}\\
  \texttt{m.valentino@sheffield.ac.uk}\quad \texttt{andre.freitas@idiap.ch}
  }

\begin{document}
\maketitle
\begin{abstract}
Thanks to their linguistic capabilities, LLMs offer an opportunity to bridge the gap between informal mathematics and formal languages through \emph{autoformalization}. However, it is still unclear how well LLMs generalize to sophisticated and naturally occurring mathematical statements. To address this gap, we investigate the task of autoformalizing real-world \emph{mathematical definitions}: a critical component of mathematical discourse. Specifically, we introduce two novel resources for autoformalization, collecting \emph{definitions} from Wikipedia (Def\_Wiki) and arXiv papers (Def\_ArXiv). We then systematically evaluate a range of LLMs, analyzing their ability to formalize definitions into Isabelle/HOL. Furthermore, we investigate strategies to enhance LLMs' performance including \emph{refinement through external feedback} from Proof Assistants, and \emph{formal definition grounding}, where we augment LLMs' formalizations through relevant contextual elements from formal mathematical libraries. Our findings reveal that definitions present a greater challenge compared to existing benchmarks, such as miniF2F. In particular, we found that LLMs still struggle with self-correction, and aligning with relevant mathematical libraries. At the same time, structured refinement methods and definition grounding strategies yield notable improvements of up to 16\% on self-correction capabilities and 43\% on the reduction of undefined errors, highlighting promising directions for enhancing LLM-based autoformalization in real-world scenarios.\footnote{Code and datasets are available at \url{https://github.com/lanzhang128/definition_autoformalization}}
\end{abstract}

\section{Introduction}

Large Language Models (LLMs) have demonstrated remarkable potential in assisting with mathematical reasoning on different downstream tasks ~\citep{wei2022chain,meadows2023generating,meadows-etal-2024-symbolic,valentino-etal-2022-textgraphs,lu-etal-2023-survey,meadows2023introduction,mishra-etal-2022-lila,ferreira-etal-2022-integer,ferreira2020premise,welleck2021naturalproofs,mishra2022numglue,petersen-etal-2023-neural}. In the context of mathematics, formal languages play a crucial role by providing a precise, logic-based framework for verifying the correctness and logical validity of mathematical statements and proofs~\citep{survey2020}. 
\begin{figure}[!t]
    \centering
    \includegraphics[width=\columnwidth]{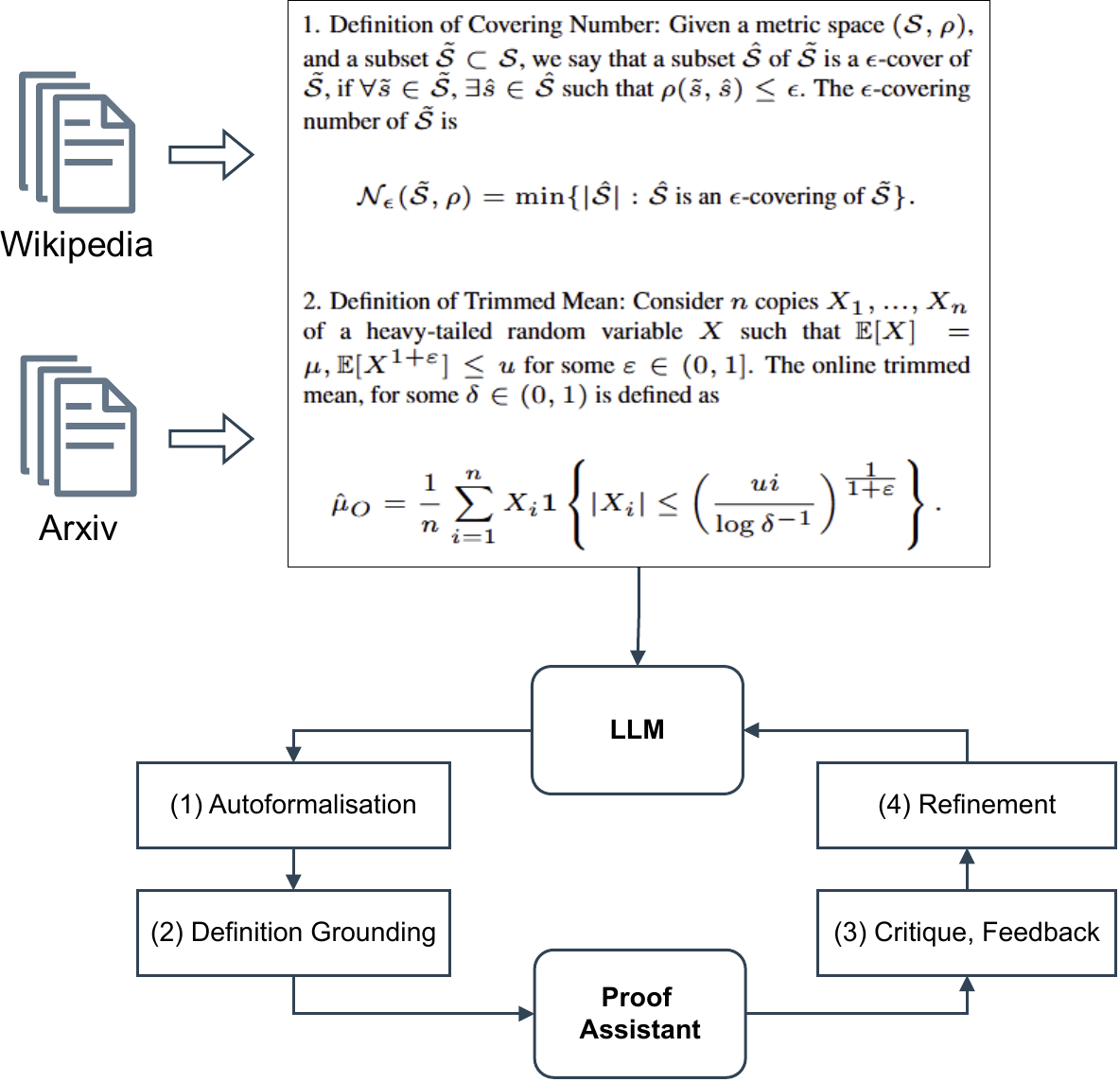}
    \caption{ \emph{Can LLMs formalize complex mathematical statements?} This paper investigates the task of translating \emph{real-world mathematical definitions} into a formal language. We introduce a new resource collecting definitions from \emph{Wikipedia} and \emph{ArXiv} papers,  exploring different strategies for autoformalization through the interaction between \emph{LLMs} and \emph{Proof Assistants}.}
    \label{fig:framework}
\end{figure}
Consequently, one promising application of LLMs is \emph{autoformalization}, the task of translating informal statements into formal languages~\citep{wu2022autoformalization}. Given their advanced linguistic and inferential capabilities, LLMs offer an opportunity to bridge the gap between informal mathematics, natural language, and machine-verifiable logic, potentially streamlining and scaling the process of formal mathematical reasoning~\citep{jiang2023draft,tarrach2024more}.

The task of autoformalization has collected increasing attention in recent years, leading to the development of benchmarks and evaluation methodologies~\citep{azerbayev2023proofnet,zhang-etal-2024-consistent,li2024autoformalize}. Despite this progress, however, existing benchmarks for autoformalization often focus on relatively simple mathematical problems, limiting our understanding of how well LLMs generalize to more sophisticated and naturally occurring mathematical statements.

To address this gap, this paper investigates the task of autoformalizing \emph{mathematical definitions}, a critical component of mathematical discourse~\citep{Moschkovich2003WhatCA}. Definitions serve as foundational building blocks in mathematical reasoning, yet they are often intricate, context-dependent, and thus difficult to formalize. Evaluating LLMs on this subset of mathematical statements, therefore, allows for assessing their ability to formally represent fine-grained mathematical concepts, highlighting persisting challenges and limitations for real-world applications.



Specifically, this paper introduces two new benchmarks for autoformalization by collecting \emph{real-world mathematical definitions} into two distinct resources:
 (1) \emph{Def\_Wiki}, including definitions extracted from Wikipedia articles, and (2) \emph{Def\_ArXiv}, including definitions collected from machine learning research papers.
Using these resources, we first evaluate LLMs in a zero-shot setting, analyzing their ability to translate definitions into Isabelle/HOL~\citep{Nipkow-Paulson-Wenzel:2002}. 

Furthermore, to address observed limitations, we investigate two key strategies to enhance performance: (1) \emph{Refinement via external feedback}, investigating the self-correction capabilities of LLMs by incorporating errors found by the supporting Proof Assistant. In particular, we show that while LLMs exhibit limited ability to refine outputs based on binary feedback (error vs. non-error), a more structured categorical refinement implemented via additional instructional constraints can improve performance. (2) \emph{Formal definition grounding}. Many mathematical definitions require references to formal objects in external mathematical libraries. To augment autoformalization from LLMs, we investigate the impact of introducing additional contextual control mechanisms, which add contextual elements from formal mathematical libraries as auxiliary premises.

Overall, our findings reveal that the proposed benchmarks present a greater challenge compared to existing autoformalization datasets, such as miniF2F~\citep{zheng2022miniff}. In particular, LLMs struggle with self-correction and particularly with incorporating relevant mathematical libraries as preambles. Proposed structured refinement methods and definition grounding strategies both deliver notable improvements, highlighting promising directions for enhancing LLM-based autoformalization in real-world scenarios.

Our contributions can be summarized as follows:
\begin{enumerate}
    \item We introduce and release two novel datasets for autoformalization: Def\_Wiki (definitions from Wikipedia) and Def\_ArXiv (definitions from research papers on arXiv), designed to assess LLMs performance on complex, real-world mathematical definitions.
    \item We perform a comprehensive error analysis on Isabelle/HOL, identifying key failures in formalizations generated by LLMs spanning across different families, including GPT-4o~\citep{openai2024gpt4o}, Llama3~\citep{grattafiori2024llama3herdmodels} and DeepSeekMath~\citep{shao2024deepseekmath}.
    \item We investigate refinement-based strategies, including structured feedback mechanisms from Proof Assistants and instruction-based categorical refinements.
    \item We explore the role of formal definition grounding, investigating how the inclusion of relevant mathematical libraries impacts the ability of LLMs to connect the formalized statements with contextual mathematical elements and relevant premises.
\end{enumerate}

\begin{table*}[!t]
    \small
    \centering
    \begin{tabular}{p{0.2\textwidth}| p{0.32\textwidth} | p{0.39\textwidth}}
        \toprule
        miniF2F & Def\_Wiki & Def\_ArXiv\\
        \midrule
        1. Suppose that $\sec x+\tan x=\frac{22}7$ and that $\csc x+\cot x=\frac mn,$ where $\frac mn$ is in lowest terms.  Find $m+n^{}_{}.$ Show that it is 044.\newline
        2. What is the sum of the two values of $x$ for which $(x+3)^2 = 121$? Show that it is -6.\newline
        3. The product of two positive whole numbers is 2005. If neither number is 1, what is the sum of the two numbers? Show that it is 406.\newline
        4. The expression $10x^2-x-24$ can be written as $(Ax-8)(Bx+3),$ where $A$ and $B$ are integers. What is $AB + B$? Show that it is 12.
        & 
        1. Definition of Rademacher Complexity: Given a set \(A\subseteq \mathbb{R}^m\), the Rademacher complexity of A is defined as follows: \[\operatorname{Rad}(A):=\frac{1}{m}\mathbb{E}_\sigma\left[\sup_{a \in A}\sum_{i=1}^m \sigma_i a_i\right]\] where \(\sigma_1, \sigma_2, \dots, \sigma_m\) are independent random variables drawn from the Rademacher distribution (i.e. \(\Pr(\sigma_i = +1) = \Pr(\sigma_i = -1) = 1/2\) for \(i=1,2,\dots,m\)), and \(a=(a_1,\dots,a_m)\).\newline 
        2. Definition of Polynomial Kernel: For degree-\(d\) polynomials, the polynomial kernel is defined as \(K(\mathbf{x},\mathbf{y}) = (\mathbf{x}^\mathsf{T} \mathbf{y} + c)^{d}\) where \(\mathbf{x}\) and \(\mathbf{y}\) are vectors of size \(n\) in the input space, i.e. vectors of features computed from training or test samples and \(c\geq 0\) is a free parameter trading off the influence of higher-order versus lower-order terms in the polynomial.
        & 
        1. Definition of Covering Number: Given a metric space $(\mathcal{S}, \rho)$, and a subset $\tilde{\mathcal{S}} \subset \mathcal{S}$, we say that a subset $\hat{\mathcal{S}}$ of $\tilde{\mathcal{S}}$ is a $\epsilon$-cover of $\tilde{\mathcal{S}}$, if $\forall \tilde{s} \in \tilde{\mathcal{S}}$, $\exists \hat{s} \in \hat{\mathcal{S}}$ such that $\rho(\tilde{s}, \hat{s}) \leq \epsilon$. The $\epsilon$-covering number of $\tilde{\mathcal{S}}$ is
        \begin{displaymath}
        \mathcal{N}_{\epsilon}(\tilde{\mathcal{S}}, \rho) = \min\{ |\hat{\mathcal{S}}|:
        \hat{\mathcal{S}} \text{ is an } \epsilon\text{-covering of } \tilde{\mathcal{S}} \} .
        \end{displaymath}\newline
        2. Definition of Trimmed Mean: Consider $n$ copies $X_1, ..., X_n$ of a heavy-tailed random variable $X$ such that $\mathbb E[X] = \mu, \mathbb E[X^{1+\varepsilon}]\leq u$ for some $\varepsilon\in(0, 1]$. The online trimmed mean, for some $\delta \in (0, 1)$ is defined as
        \begin{equation*}
        \hat{\mu}_O = \frac{1}{n}\sum_{i=1}^n X_i \bm{1}\left\{|X_i| \leq \left(\frac{ui}{\log \delta^{-1}}\right)^{\frac{1}{1+\varepsilon}}\right\}.
        \end{equation*}\\
        \bottomrule
    \end{tabular}
    \caption{Examples of instances from Def\_Wiki and Def\_ArXiv and comparison with miniF2F.}
    \label{tab:data_example}
\end{table*}


\section{Autoformalization with LLMs}

The task of autoformalization can be defined as a transformation function from natural language and LaTeX symbols $\mathcal{S}$ to a formal language $\mathcal{F}$, $f: \mathcal{S} \to \mathcal{F}$, such that for every informal mathematical statement $s \in \mathcal{S}$, there exists a formal mathematical statement $\phi \in \mathcal{F}$ where $f(s) = \phi$~\citep{zhang-etal-2024-consistent}. Autoformalization via LLMs reifies the transformation function as:
\[
f(s)=\text{LLM}(p_\text{auto},\{(s_i,\phi_i)\},s),
\]
\noindent where $p_\text{auto}$ is a prompt for autoformalization and $\{(s_i,\phi_i)\}$ is an optional set of exemplars.

\subsection{Limitations of Existing Benchmarks}

Naturally occurring mathematical statements typically involve complex and abstract mathematical concepts. However, the statements in existing datasets, such as miniF2F~\citep{zheng2022miniff}, primarily consist of basic arithmetic operations and elementary mathematical objects, such as integers, fractions, and real numbers (as shown in Table~\ref{tab:data_example}). Such mathematical objects are relatively simple compared to the complex and abstract concepts found in naturally occurring mathematical statements and scientific papers, which may involve higher-level structures like vectors, matrices, and probability. The operations are also limited to simple arithmetic, such as addition, subtraction, multiplication, division, and exponentiation. Studying autoformalization on such datasets, therefore, does not necessarily reflect the challenges of autoformalization in realistic scenarios. However, few benchmarks focus on how to construct complex mathematical statements. Our work aims to address this gap.

\subsection{Real-World Mathematical Definitions}
Since extracting high-quality definition statements from general mathematical corpora requires careful curation, we propose a systematic data creation process that balances complexity and diversity. We begin by classifying real-world definitions into two categories: (i) common definitions, which are presented in a global context where the necessary preliminaries are \textit{implicit} and \textit{relatively general}, and (ii) specialized definitions, which are typically situated in a local context and rely on \textit{explicit} and \textit{specific} preliminaries. Both types present distinct challenges for autoformalization. In the former, the model must infer implicit preliminaries and connect them with existing formal constructs. In the latter, the model must not only formalize the definition but also the associated preliminaries. We ground these two types of definitions in two sources: Wikipedia for common definitions (Def\_Wiki) and Arxiv Papers for specialized definitions (Def\_ArXiv), as definitions from these two sources are human-written, naturally occurring and likely to have already been validated.

To construct a representative instance for Def\_Wiki and Def\_Arxiv, we focus on definitions from the field of machine learning, as this domain offers a sufficient number of novel, diverse, and relatively unformalized definitions. Moreover, the task of formalizing such definitions holds practical value for the AI community. The detailed operational steps are provided in \textit{Appendix~\ref{app:data}}. A quality check was performed to ensure that the selected definitions exhibit high diversity and present complementary challenges. This process yielded 56 definitions for Def\_Wiki and 30 definitions for Def\_Arxiv. This number is expected, as novel definitions in the real world are relatively scarce compared to example questions in mathematics (such as those in miniF2F), which are often closer in nature to synthetically generated content. Although the resulting datasets are relatively small in scale, they are sufficient to expose the core challenges of autoformalization in real-world scenarios. The size of the datasets is not the primary factor determining the significance, transferability, or robustness of the findings. Exploring additional scientific domains and expanding the dataset further are promising directions for future work.

We compare the quality of miniF2F with the target definition datasets. MiniF2F is significantly less abstract, complex and diverse, as intuitively shown in the randomly chosen examples in Table~\ref{tab:data_example}. The data properties are summarized in Table~\ref{tab:data_property} in \textit{Appendix}. Definition datasets exhibit higher means for the number of tokens, mathematical objects, and formulae per example, indicating that they are significantly more complex. Additionally, definition datasets have higher standard deviations, suggesting greater diversity among samples. 


The proposed benchmarks contain only definitions in LaTeX format. We did not include ground-truth formal code for the following reasons: 1. Including such code could increase the risk of data leakage, as ground-truth formalizations from publicly available datasets may have been seen by LLMs, whose training data is not fully disclosed, making it difficult to determine whether future improvements genuinely address the challenges posed by our benchmark or simply reflect prior exposure. 2. A single mathematical statement can have multiple correct formalizations. An autoformalized output that differs from the reference does not necessarily indicate incorrect formalization. 3. The main purpose of ground-truth formal code is to evaluate autoformalization. However, the syntactic correctness of formalized code can be rigorously and automatically verified using theorem provers~\citep{zhang-etal-2024-consistent}, while semantic consistency can potentially be assessed in a reference-free manner via LLM-as-Judges~\citep{zhang2025goldstandardsepistemicensemble}. Moreover, manual inspection of autoformalized code does not require ground-truth formalizations.

\begin{table*}[!t]
    \centering
    \small
    \begin{tabular}{l l c c| c c c c c}
        \toprule
        Prompt Strategy & Model & Pass$\uparrow$ & FEO$\uparrow$ & TRO$\downarrow$ & IVI$\downarrow$  & SYN$\downarrow$ & UDF$\downarrow$ & TUF$\downarrow$\\
        \midrule
        \multicolumn{9}{l}{\textbf{miniF2F-Test}}\\
        \midrule
        ZS & DeepSeekMath-7B & 3.28 & 12.79 & 18.44 & \textbf{0.00} & 50.00 & 14.34 & 9.43\\
        ZS + Binary & & 2.05 & 6.73 & 2.46 & \textbf{0.00} & 79.91 & \textbf{5.33} & \textbf{2.05}\\
        ZS & Llama3-8B & 4.92 & 20.70 & 4.51 & 0.41 & 29.51 & 38.52 & 18.85\\
        ZS + Binary & & 3.69 & 20.52 & 3.28 & 0.41 & 33.20 & 39.75 & 20.49\\
        ZS & GPT-4o & 25.41 & 48.90 & \textbf{1.23} & 1.23 & \textbf{6.15} & 23.77 & 7.38\\
        ZS + Binary & & \textbf{29.10} & \textbf{53.90} & 2.05 & 1.23 & \textbf{6.15} & 21.72 & 8.20\\
        \midrule
        \multicolumn{9}{l}{\textbf{Def\_Wiki-Test}}\\
        \midrule
        ZS & DeepSeekMath-7B & 10.87 & 17.75 & 34.78 & 2.17 & 30.43 & 26.09 & \textbf{2.17}\\
        ZS + Binary & & 6.52 & 7.73 & 8.70 & \textbf{0.00} & 69.57 & \textbf{21.74} & \textbf{2.17}\\
        ZS & Llama3-8B & 0.00 & 2.80 & \textbf{0.00} & 23.91 & 56.52 & 32.61 & 4.35\\
        ZS + Binary & & 2.17 & 3.71 & \textbf{0.00} & 26.09 & 52.17 & 30.43 & \textbf{2.17}\\
        ZS & GPT-4o & 10.87 & 16.12 & 8.70 & 8.70 & 19.57 & 50.00 & 13.04\\
        ZS + Binary & & \textbf{13.04} & \textbf{18.30} & 8.70 & 6.52 & \textbf{17.39} & 50.00 & 13.04\\
        \midrule
        \multicolumn{9}{l}{\textbf{Def\_ArXiv}}\\
        \midrule
        ZS & DeepSeekMath-7B & 13.33 & 14.69 & 16.67 & \textbf{0.00} & 40.00 & 36.67 & 13.33\\
        ZS + Binary & & 3.33 & 3.33 & 6.67 & \textbf{0.00} & 66.67 & \textbf{33.33} & \textbf{3.33}\\
        ZS & Llama3-8B & 0.00 & 2.67 & \textbf{0.00} & 13.33 & 70.00 & 40.00 & 6.67\\
        ZS + Binary & & 3.33 & 5.83 & \textbf{0.00} & 20.00 & 60.00 & \textbf{33.33} & 6.67\\
        ZS & GPT-4o & 13.33 & 19.30 & \textbf{0.00} & \textbf{0.00} & 40.00 & 56.66 & 6.67\\
        ZS + Binary & GPT-4o & \textbf{16.67} & \textbf{24.30} & \textbf{0.00} & \textbf{0.00} & \textbf{33.33} & 53.33 & 6.67\\
        \bottomrule
    \end{tabular}
    \caption{Autoformalization results. Prompt strategies include: (\textbf{ZS}): zero-shot prompting; (\textbf{ZS + Binary}): refinement given (ZS) formalized code and binary syntactic correctness state. Pass rate (\textbf{Pass}), the place of first error occurrence in the main body of the code (\textbf{FEO}), and percentage of occurrence of each error category are recorded here. Errors in each error category are: (\textbf{TRO}): Time Run-Out for checking; (\textbf{IVI}): Fake Non-Existent Theory, Invalid structural format; (\textbf{SYN}): Inner syntax error, Outer syntax error, Inner lexical error, Malformed command syntax, Bad name, Bad number of arguments for type constructor, Extra free type variable(s); (\textbf{UDF}): Undefined type names, Undeclared class, Undefined locale, No type arity list, Extra variables on rhs; (\textbf{TUF}) Type unification failed.}
    \label{tab:zs}
\end{table*}

\section{Empirical Evaluation}

\paragraph{Empirical Setup.} Isabelle/HOL was chosen as the representative formal language due to its widespread adoption within the formal mathematics community and its ability to provide specific information about the types of errors encountered. Moreover, Isabelle employs a declarative proof language that is closer to structured natural language, making it well-suited for exploring autoformalization of complex natural language statements. We prioritize an in-depth analysis of a single formal language over a broad comparison across multiple languages. We believe this approach lays the foundation for future work aimed at exploring the behavior of LLMs in alternative formal systems. We evaluate three LLMs with different features: DeepSeekMath-7B~\citep{shao2024deepseekmath}, Llama3-8B~\citep{grattafiori2024llama3herdmodels} and GPT-4o~\citep{openai2024gpt4o}. DeepSeekMath-7B is an open-sourced LLM trained specifically for mathematics. As a smaller model, it has demonstrated comparable mathematical reasoning performance as in GPT-4~\citep{openai2024gpt4}, and strong few-shot autoformalization performance on miniF2F with Isabelle. This superiority makes it a good representative of smaller but specialized LLMs. LLama3-8B is a smaller open-sourced foundation LLM with no specific emphasis on math. GPT-4o is widely acknowledged as one of the state-of-the-art LLMs. For reproducibility, greedy decoding is used for generation in all settings. 

\paragraph{Evaluation Metrics.} The success rate of passing the check by the Isabelle Proof Assistant across the tested dataset is used as the first metric. We assume that a formalized code instance with the first error occurring later in the code reflects, as a proxy, the level of autoformalization. Thus, we evaluate such by calculating the proportion of correct lines (up to the first error) within the main body of the code. For syntactically correct instances, this value is equal to 1. To better monitor the occurrence of errors, we group them into three categories: Syntax Errors (SYN), Undefined Item Errors (UDF), and Type Unification Failed Errors (TUF). For each category, we calculate the percentage of incorrect formalized codes caused by errors in that category.

\subsection{Zero-Shot Prompting \& Binary Refinement}

In order to understand the challenges in autoformalizing mathematical definitions with LLMs, we perform a preliminary analysis on miniF2F~\citep{zheng2022miniff}, Def\_Wiki and Def\_ArXiv using zero-shot prompting (ZS) and binary refinement. With binary refinement, we aim to assess the capabilities of LLMs for error correction by providing them with the formal code generated via ZS, along with the syntactic correctness evaluated using the proof assistant (i.e., ``correct'', ``incorrect''). From the results reported in Table~\ref{tab:zs}, we can derive the following observations:

\paragraph{Def\_Wiki and Def\_ArXiv are significantly more challenging than miniF2F.} When performing autoformalization on Def\_Wiki and Def\_ArXiv, GPT-4o achieves a significantly lower success rates (-13.78\% on average) and FEO (-31.90\% on average) compared to results on miniF2F-Test.

\paragraph{LLMs can provide false preambles when performing autoformalization.} In Table~\ref{tab:zs}, the percentage of Invalid Inputs errors (IVI) can be non-zero. Errors in this category are caused by either non-existent preambles or invalid file formats in structure. For Llama3-8B the latter is more common whereas for GPT-4o, we observe that the dominant cause is the generation of non-existent preambles, showing that GPT-4o do not perfectly generalize in recognizing the names of preambles.

\begin{table*}[!t]
    \small
    \centering
    \begin{tabular}{p{0.08\textwidth} p{0.85\textwidth}}
        \toprule
        Category & Reasons\\
        \midrule
        SYN & 
        1. \textbf{Invalid Symbol Format.} Isabelle uses symbols like ``\textbackslash<Rightarrow>'' to represent ``\textbackslash Rightarrow ($\Rightarrow$)'' in LaTeX. GPT-4o does not strictly follow this behaviour. A symbol in its formalized code starting with ``\textbackslash<'' can miss ``>'' at the end so that the relevant symbol is not valid.\newline
        2. \textbf{Confusion of Mapping between LaTeX Mathematical Symbols and Isabelle Symbols.} Not all natural language symbols in LaTeX have a similar corresponding version in Isabelle symbols. In natural language mathematics we use different mathematical fonts such as ``\textbackslash mathcal ($\mathcal{A}$)'' to distinguish items. Isabelle uses ``\textbackslash<A>'' to represent this LaTeX symbol. However, GPT-4o would pretend the existence of a symbol named \textbackslash<mathcal> and use it for autoformalization.\newline 
        3. \textbf{Unaware of Name Conflict.} Some keywords such as ``instance'' are reserved by Isabelle/HOL and they cannot be used as the name of a new item.\newline
        4. \textbf{Incorrect Stylistic Usage of Symbols or Operators.} Some symbols or operators require specific usage which is not in the same style as in natural language. The incorrect usage of them in formalized code generated by GPT-4o can lead to syntax errors.\\
        \midrule
        UDF & 1. \textbf{Items not defined.} Formalization requires every mentioned item  to be clearly defined in the local context or preambles. During autoformalization, GPT-4o could refer to items that are not defined in both sources.\\
        \midrule
        TUF & 1. \textbf{Mismatch between Types in Definition and Types in Actual Usage.} There are some operators or functions which have been clearly defined about the types of their operands or parameters. When using these operators or functions, the types of actual operands or parameters need to match the types in the definitions exactly. GPT-4o would produce mismatched types in the formalized codes and introduce TUF errors.\\
        \bottomrule
    \end{tabular}
    \caption{Reasons of failure in each error category during autoformalization with GPT-4o.}
    \label{tab:failure}
\end{table*}

\paragraph{Specialized smaller models can reach the same level of success rate as larger LLMs.} As a model designed specifically for mathematics, DeepSeekMath-7B can achieve a similar success rate as GPT-4o. Although Llama3-8B has a larger model size, its generalization ability on definitions is limited. Additionally, DeepSeekMath-7B exhibits a lower percentage of undefined type names errors (UDF). However, one disadvantage of the specialized model is that its formalizations have a higher percentage of time run-out issues (TRO). This is likely caused by the bias introduced during the fine-tuning phase on theorem proving which can lead the model to generate unsolicited proofs.

\paragraph{Small LLMs possess limited binary self-correction capabilities.} With binary refinement, GPT-4o produces formal codes with a higher success rate on all three datasets, whereas for DeepSeekMath-7B this mechanism leads to a performance decrease. LLama3-8B also fails to self-correct its autoformalization results on miniF2F. This behavior suggests that self-refinement exceeds the capabilities of smaller LLMs.

\subsubsection{Error Analysis \& Interventions}

To understand potential interventions for improving autoformalization, we qualitatively analyze error patterns on the development set of Def\_Wiki. Our analysis is based on the results obtained via GPT-4o, given its better performances on ZS and binary refinement. The main reasons for failure identified through our analysis are summarized in Table~\ref{tab:failure}, with additional examples reported in \textit{Appendix}.

We observe that syntactic errors (SYN) exhibit the most variety, suggesting that GPT-4o may struggle to follow syntactic rules in Isabelle/HOL if not explicitly instructed. Type unification errors (TUF) suggest that GPT-4o may struggle with the exact usage of defined Isabelle items. To improve these issues, we investigate a \textbf{Categorical Refinement} (CR) method. CR involves designing specific additive instructions that constraint the behaviors leading to errors identified in the qualitative analysis.

Similarly, for syntactic errors (SYN), causes 1, 2, and 3 in Table~\ref{tab:failure} can be addressed with rule-based algorithms that refine formal codes at the symbolic level (\textbf{Symbolic Refinement}, SR). Undefined errors (UDF), on the other hand, indicate that although GPT-4o can refer to external formal mathematical items, it remains unaware of the location or existence of relevant libraries. To alleviate UDF errors, we propose the process of \textbf{Formal Definition Grounding} (FDG): linking mathematical objects mentioned in natural language statements to their formal definitions in formal libraries, and incorporating this information as contextual elements for formalizations.

\subsection{Categorical Refinement}
\begin{figure*}[!t]
    \centering
    \begin{subfigure}{0.35\textwidth} 
      \centering
      \includegraphics[width=\textwidth]{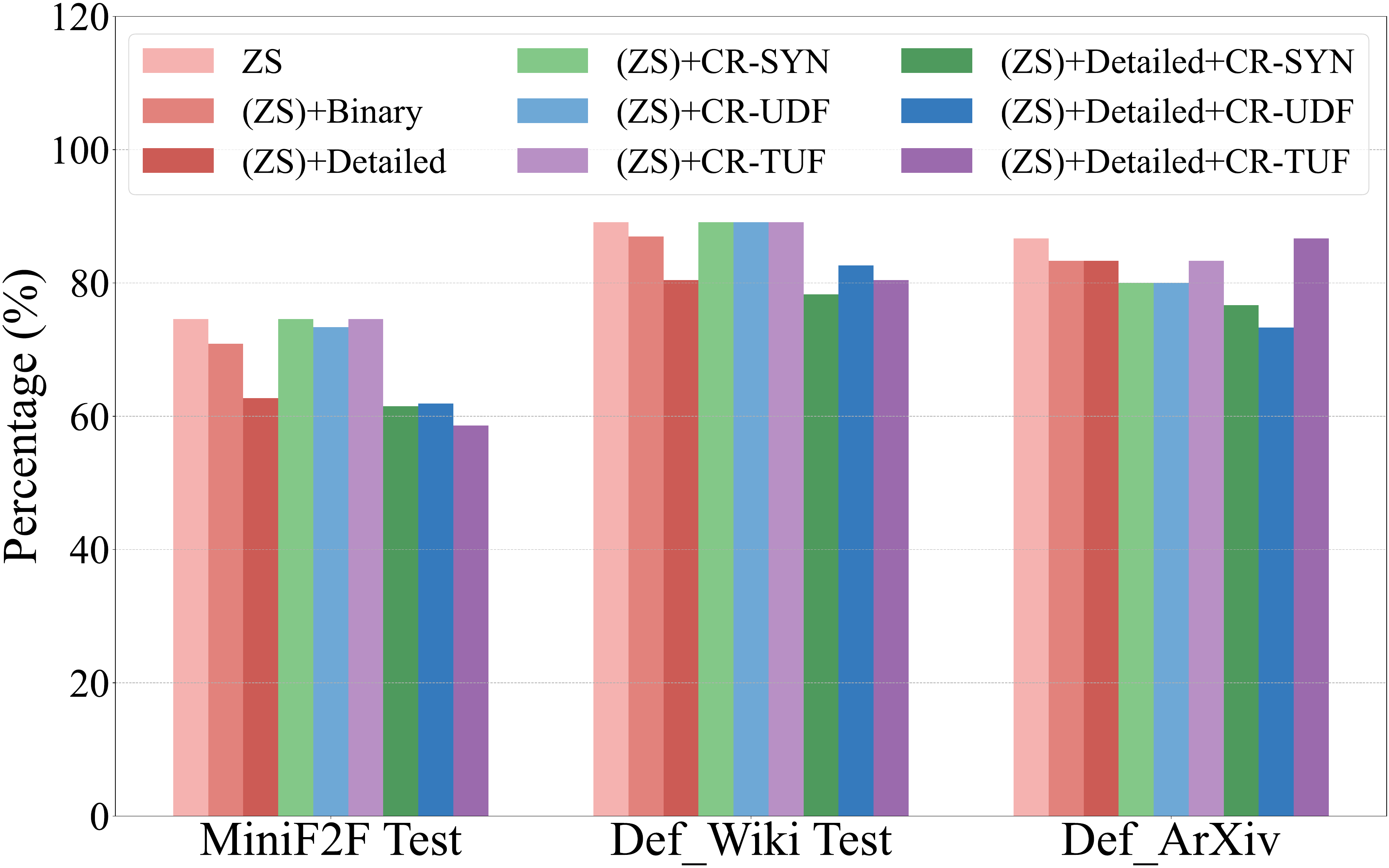}
      \caption{Overall Error Rate}
      \label{fig:success}
    \end{subfigure}
    \begin{subfigure}{0.35\textwidth} 
      \centering
      \includegraphics[width=\textwidth]{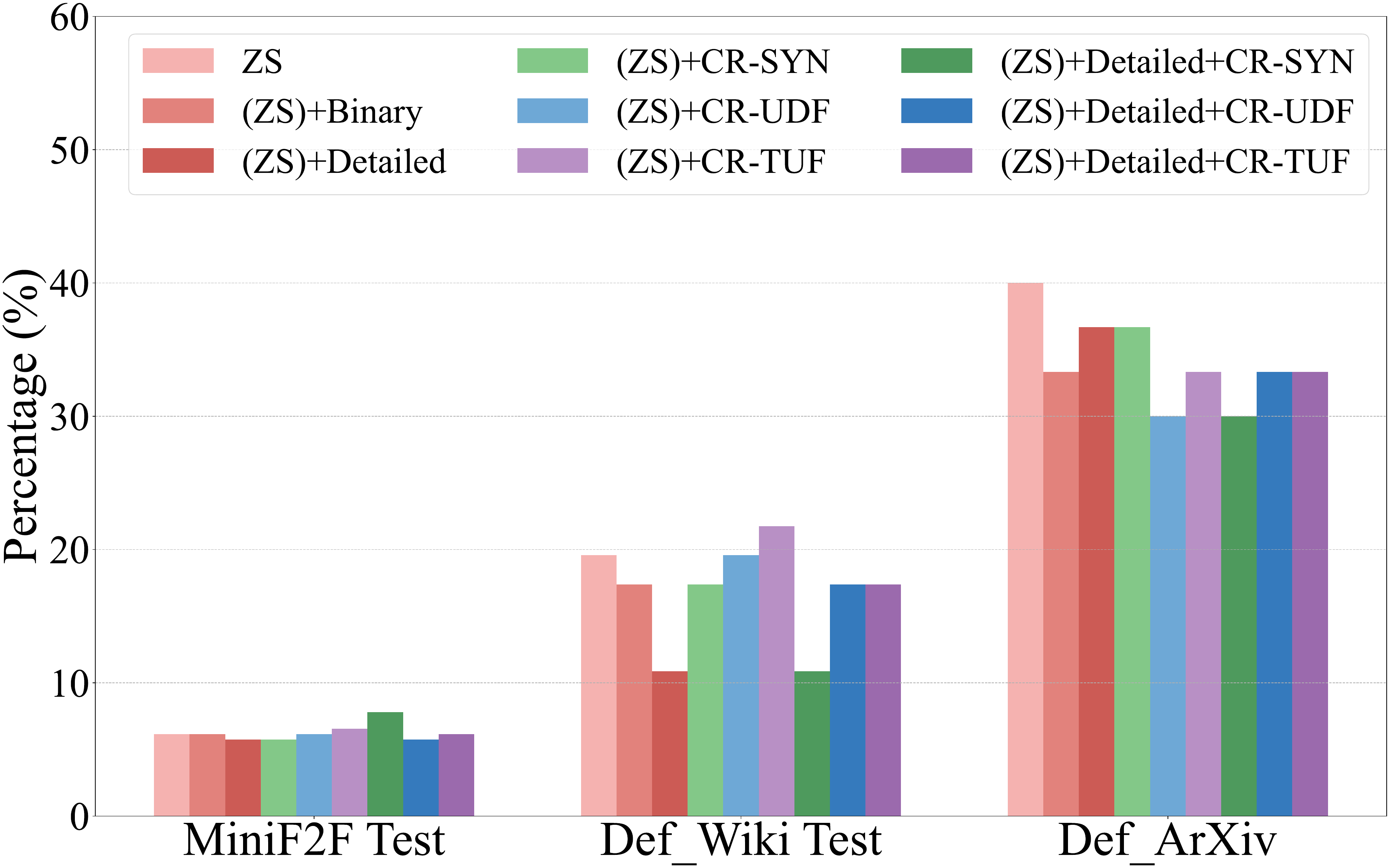}
      \caption{SYN Error Rate}
      \label{fig:syn}
    \end{subfigure}
     \begin{subfigure}{0.35\textwidth} 
      \centering
      \includegraphics[width=\textwidth]{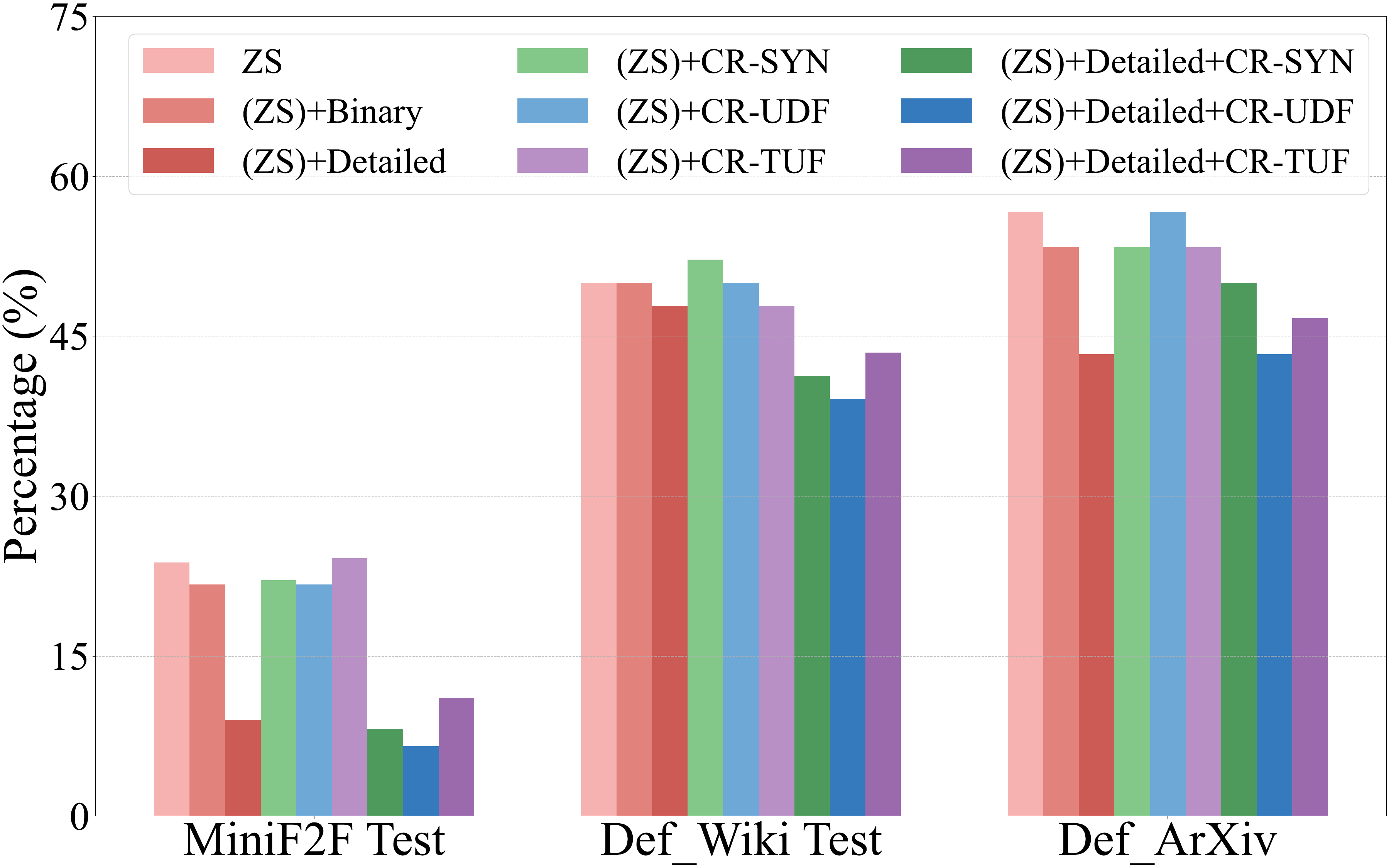}
      \caption{UDF Error Rate}
      \label{fig:udf}
    \end{subfigure}
    \begin{subfigure}{0.35\textwidth} 
      \centering
      \includegraphics[width=\textwidth]{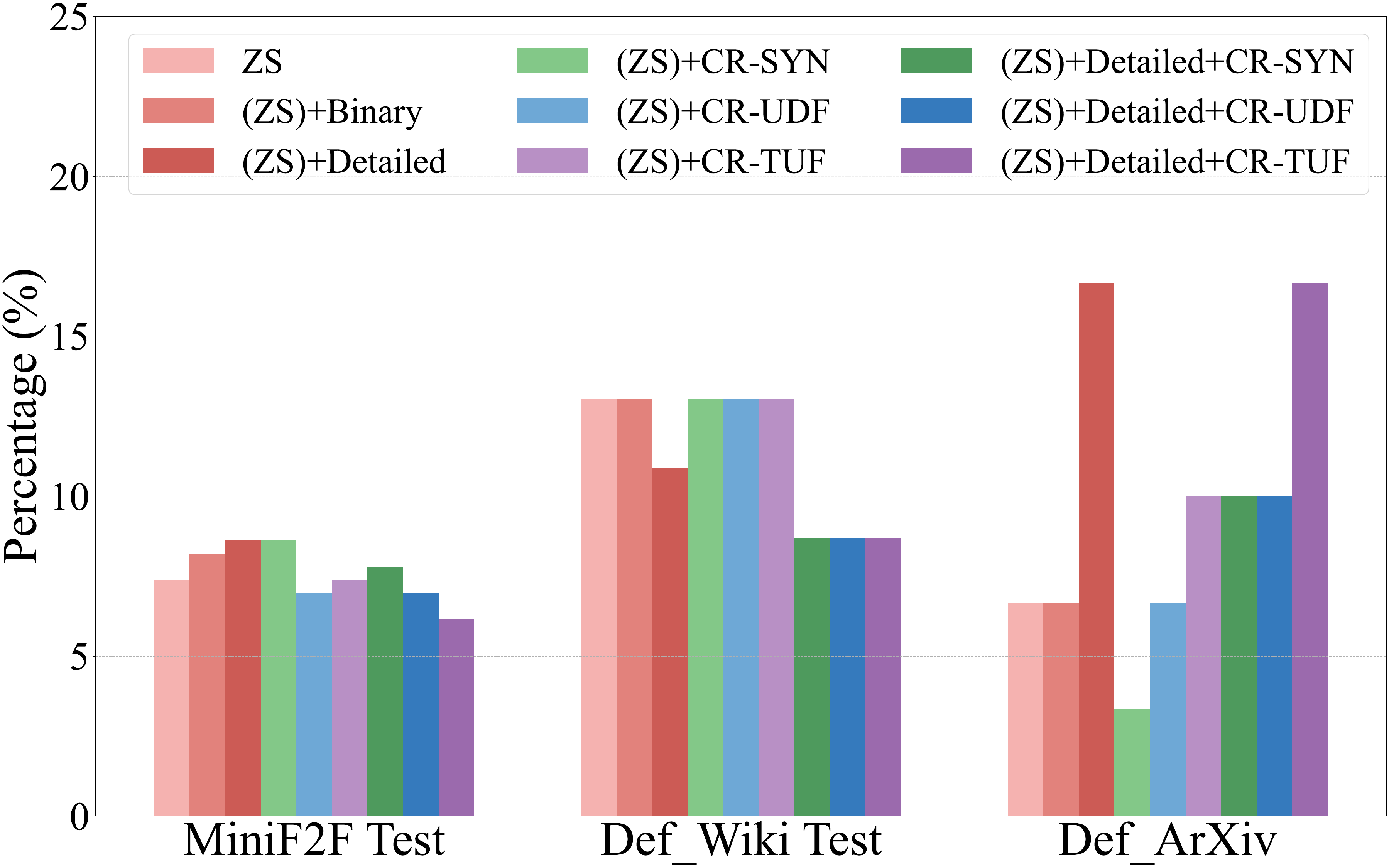}
      \caption{TUF Error Rate}
      \label{fig:tuf}
    \end{subfigure}
    \caption{Error rates of different refinement methods on GPT-4o. Variants include: (\textbf{ZS}): zero-shot autoformalization; (\textbf{(ZS)+Binary}): binary refinement on (zero-shot) formal codes; (\textbf{(ZS)+Detailed}): detailed refinement on (zero-shot) formal codes; (\textbf{(ZS)+CR-SYN/UDF/TUF}): plain refinement on (zero-shot) formal codes with SYN/UDF/TUF categorical refinement instructions; (\textbf{(ZS)+Detailed+CR-SYN/UDF/TUF}): detailed refinement on (zero-shot) formal codes with SYN/UDF/TUF categorical refinement instructions.}
    \label{fig:ref}
\end{figure*}

In order to better understand the refinement capabilities of GPT-4o, we investigate a set of error correction strategies: (i) Plain: provide LLMs with previously generated formal codes; (ii) Binary: additionally, provide LLMs with the correctness status of the formal code; (iii) Detailed: instead of just the binary correctness status, provide LLMs with the details of type, message, and line location of individual errors in the code.

In addition, to evaluate categorical refinement, we design specific instructions for each category of errors based on our qualitative analysis (Table~\ref{tab:failure}). We report the error rate results of different refinement methods on GPT-4o in bar charts in Figure~\ref{fig:ref}. All prompts used for categorical refinement and additional empirical results are provided in \textit{Appendix}.

\paragraph{Providing LLMs with more information about individual errors is more effective than simply indicating binary correctness.} As shown in Figure~\ref{fig:success}, both binary and detailed refinements can reduce the overall error rate across all the datasets, with detailed refinement fixing more errors on miniF2F-Test and Def\_Wiki-Test. For SYN errors, although there is no clear trend indicating that one refinement outperforms the other, both refinements lead to a lower error rate compared to zero-shot autoformalization. Detailed refinement also decreases the percentage of UDF errors as shown in Figure~\ref{fig:udf}. These performance gains suggest that detailed refinement improves the quality of autoformalized codes. For TUF errors, applying both refinements does not consistently result in a lower error rate, indicating that errors in this category are more difficult for LLMs to fix.

\paragraph{Categorical refinement reduces error rates.} As shown in Figure~\ref{fig:success}, across all datasets, the refinement method that achieves the lowest overall error rate incorporates one of the instructions for categorical refinement, highlighting the efficacy of this mechanism. However, when categorical refinement is applied without error details, such improvements do not occur. We hypothesize that this is because categorical instructions serve as constraints, making it more difficult for the target LLM to follow them without more detailed error information for individual instances. Once such information is provided, the LLM receives sufficient information to adhere to the categorical refinement instructions.

\paragraph{Categorical refinement can effectively reduce errors for specific categories.} As shown in Figure~\ref{fig:syn}, the method with the lowest SYN error rate on miniF2F-Test is plain refinement with SYN categorical refinement instructions, whereas on the other two datasets the best performing method is SYN categorical refinement with error details. In Figure~\ref{fig:udf}, UDF categorical refinement with error details also leads to the lowest UDF error rate on all three datasets. Similarly in Figure~\ref{fig:tuf}, TUF categorical refinement with error details achieves the lowest TUF error rate on two out of the three datasets. These results collectively demonstrate the effectiveness of the categorical refinement as a control mechanism for autoformalization. The only exception is TUF errors within the Def\_ArXiv dataset, which again highlights the difficulty of fixing TUF errors.

\begin{figure*}[!t]
    \centering
    \begin{subfigure}{0.25\textwidth} 
      \centering
      \includegraphics[width=\textwidth]{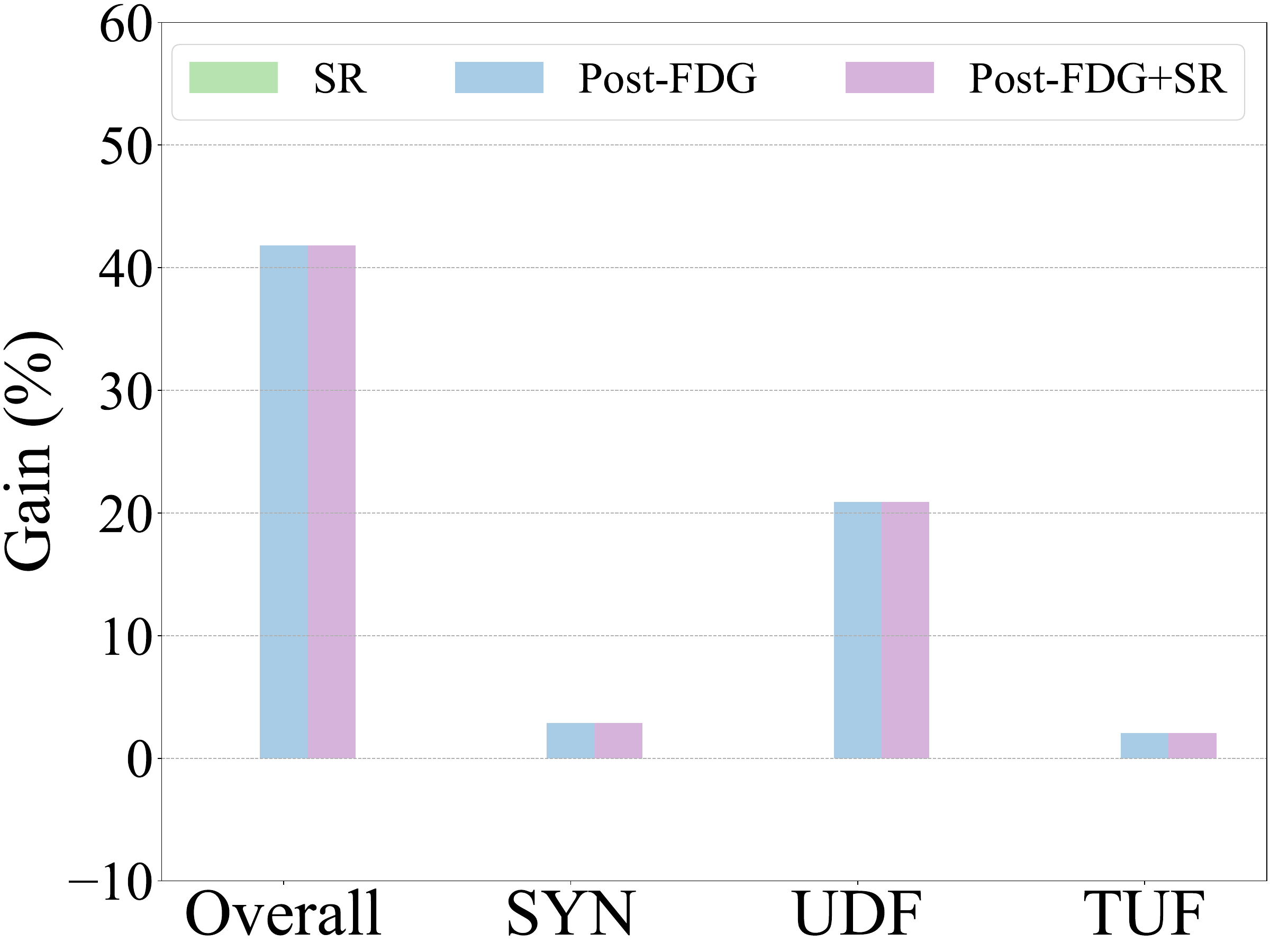}
      \caption{ZS on miniF2F}
      \label{fig:miniF2F_zs}
    \end{subfigure}
    \begin{subfigure}{0.25\textwidth} 
      \centering
      \includegraphics[width=\textwidth]{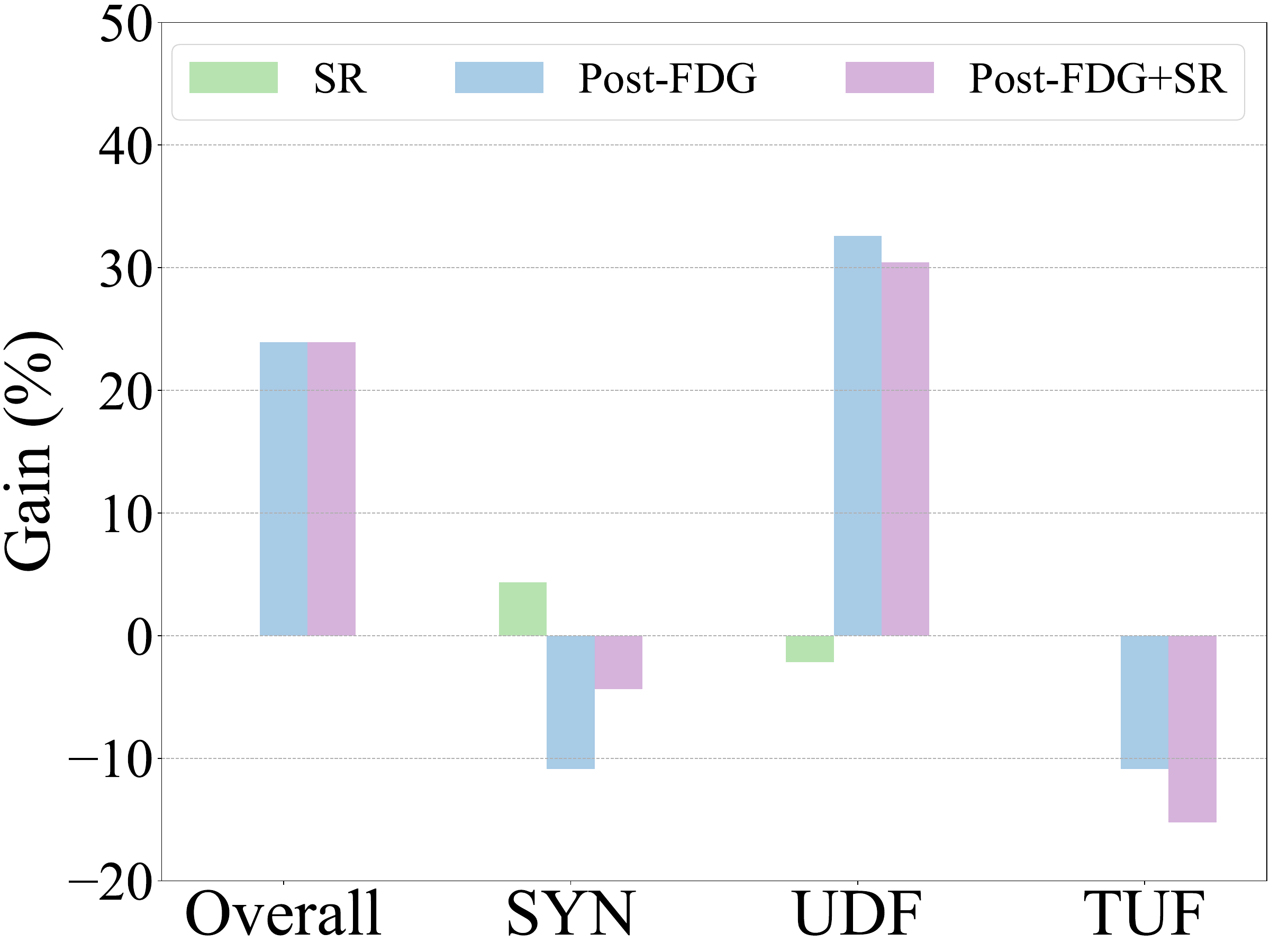}
      \caption{ZS on Def\_Wiki}
      \label{fig:wiki_zs}
    \end{subfigure}
    \begin{subfigure}{0.25\textwidth} 
      \centering
      \includegraphics[width=\textwidth]{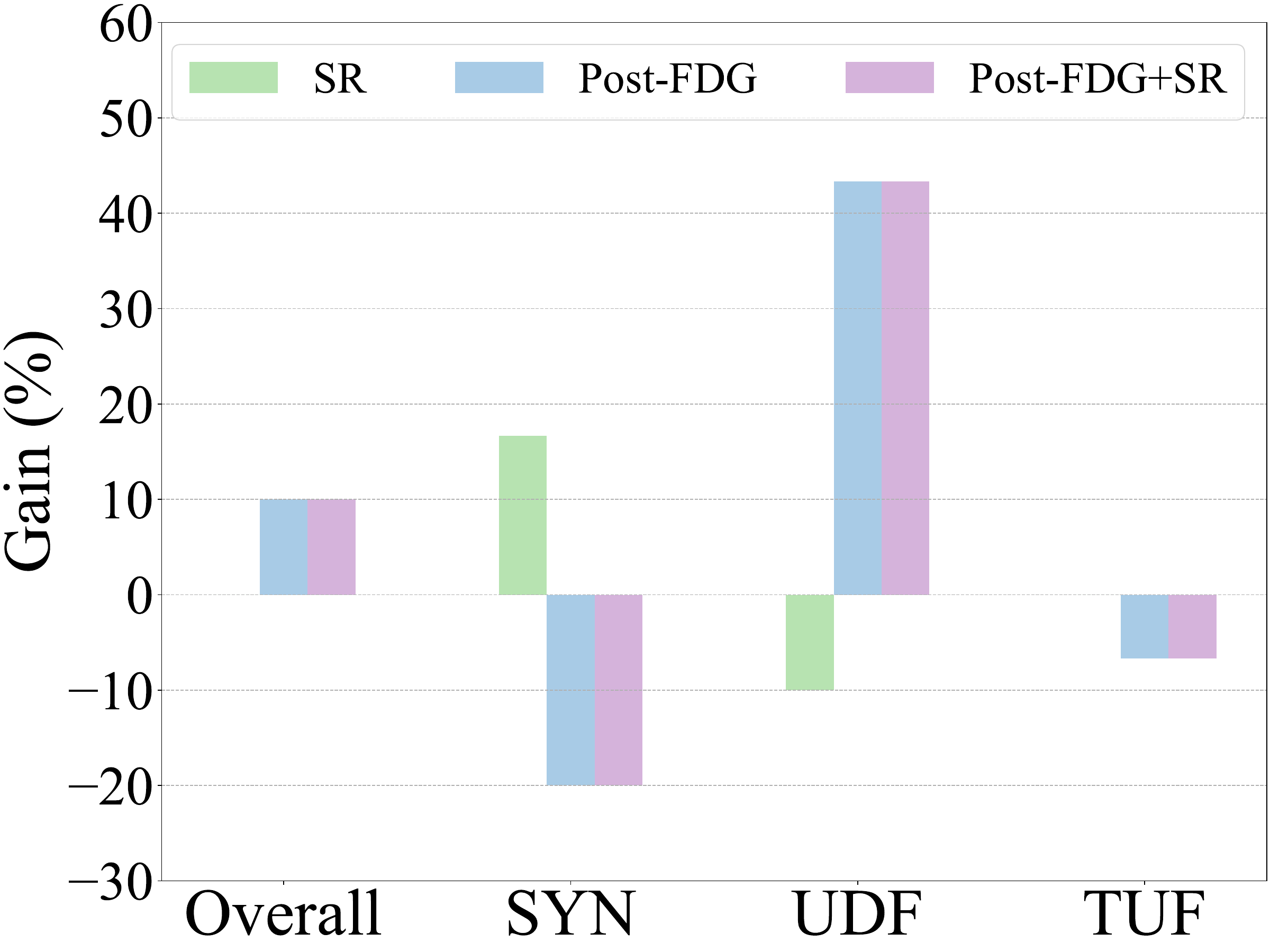}
      \caption{ZS on Def\_ArXiv}
      \label{fig:arxiv_zs}
    \end{subfigure}
    \begin{subfigure}{0.25\textwidth} 
      \centering
      \includegraphics[width=\textwidth]{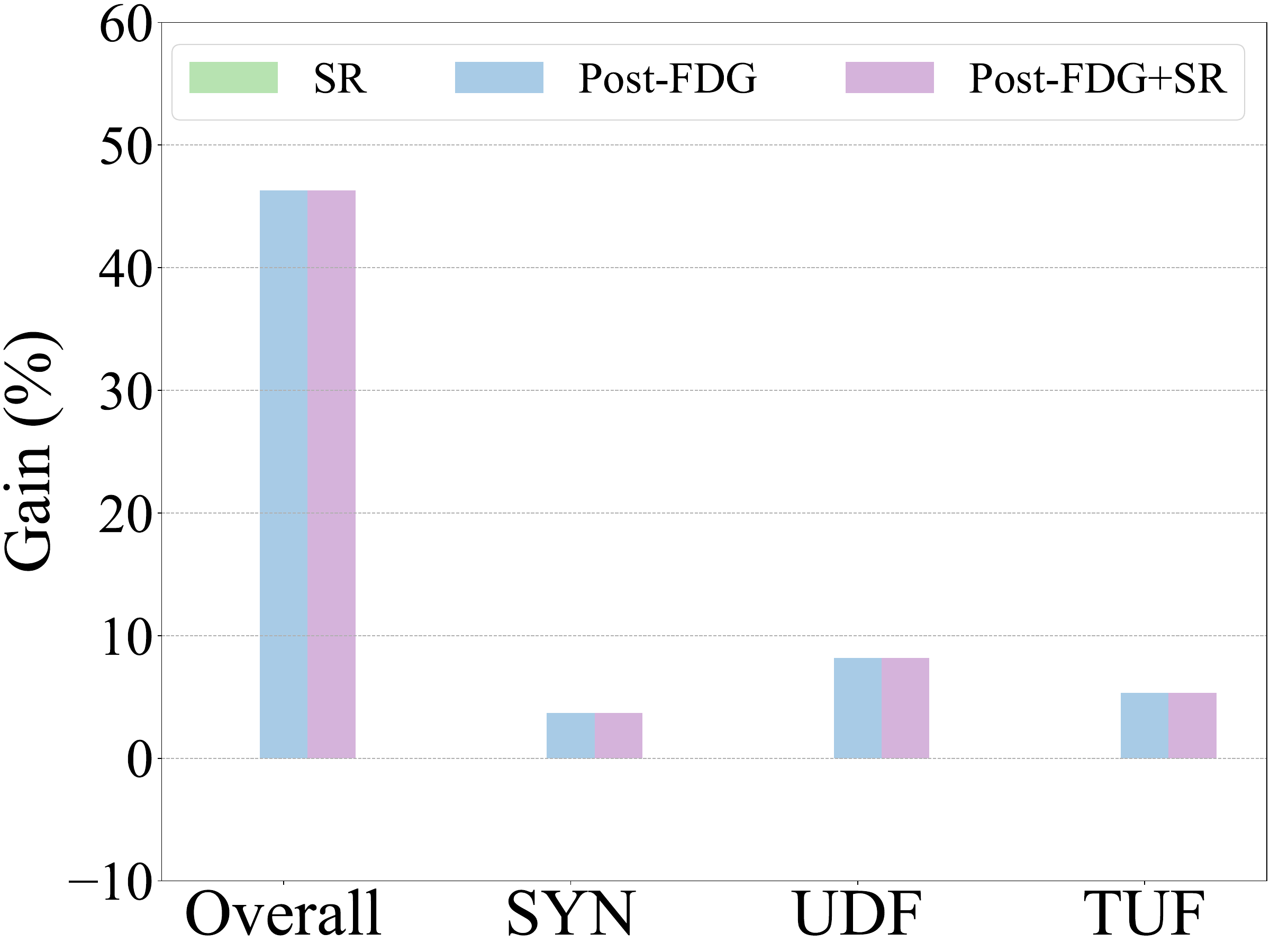}
      \caption{ZS+Detailed on miniF2F}
      \label{fig:miniF2F_zs_det}
    \end{subfigure}
    \begin{subfigure}{0.25\textwidth} 
      \centering
     \includegraphics[width=\textwidth]{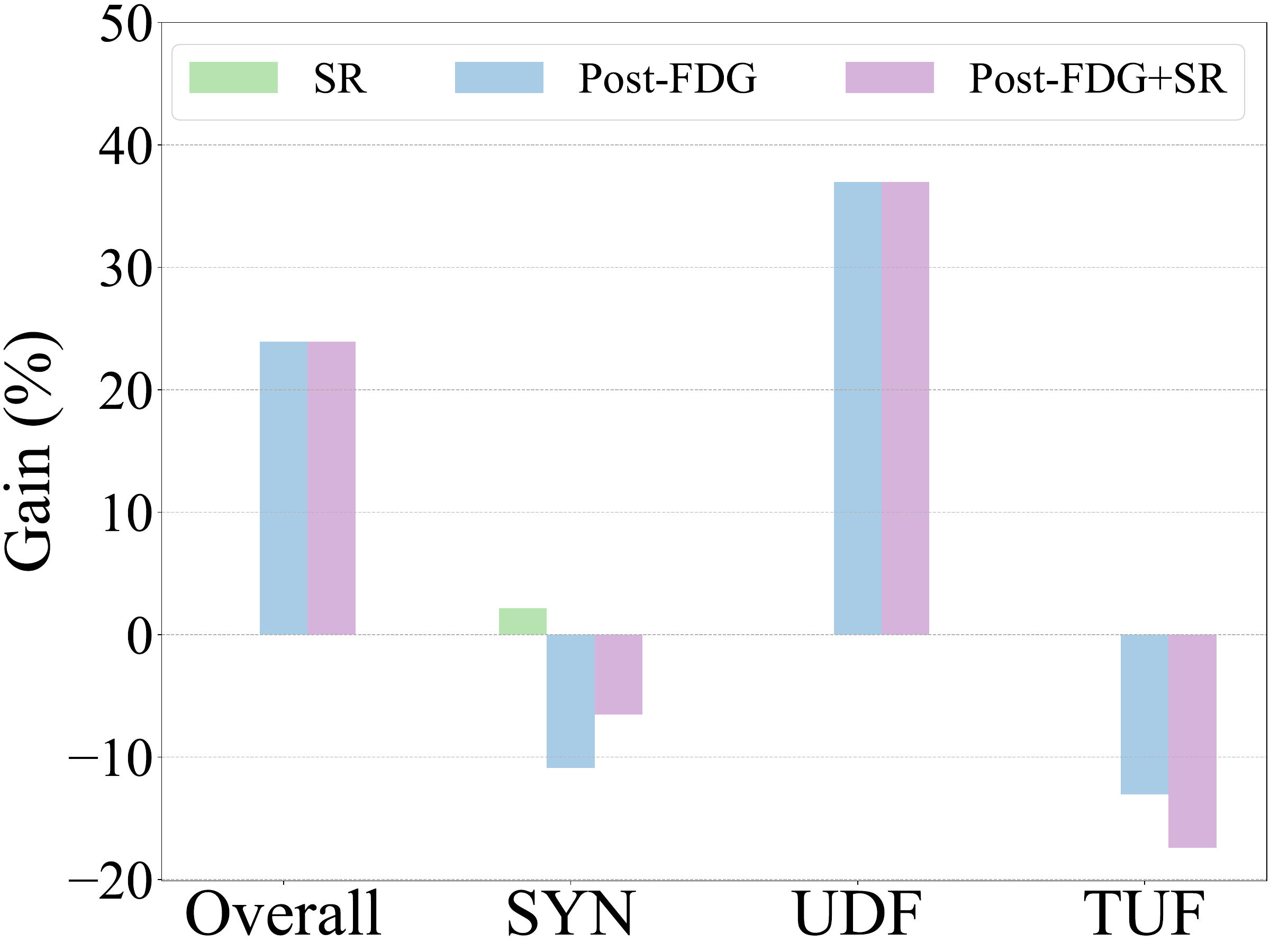}
      \caption{ZS+Detailed on Def\_Wiki}
      \label{fig:wiki_zs_det}
    \end{subfigure}
    \begin{subfigure}{0.25\textwidth} 
      \centering
    \includegraphics[width=\textwidth]{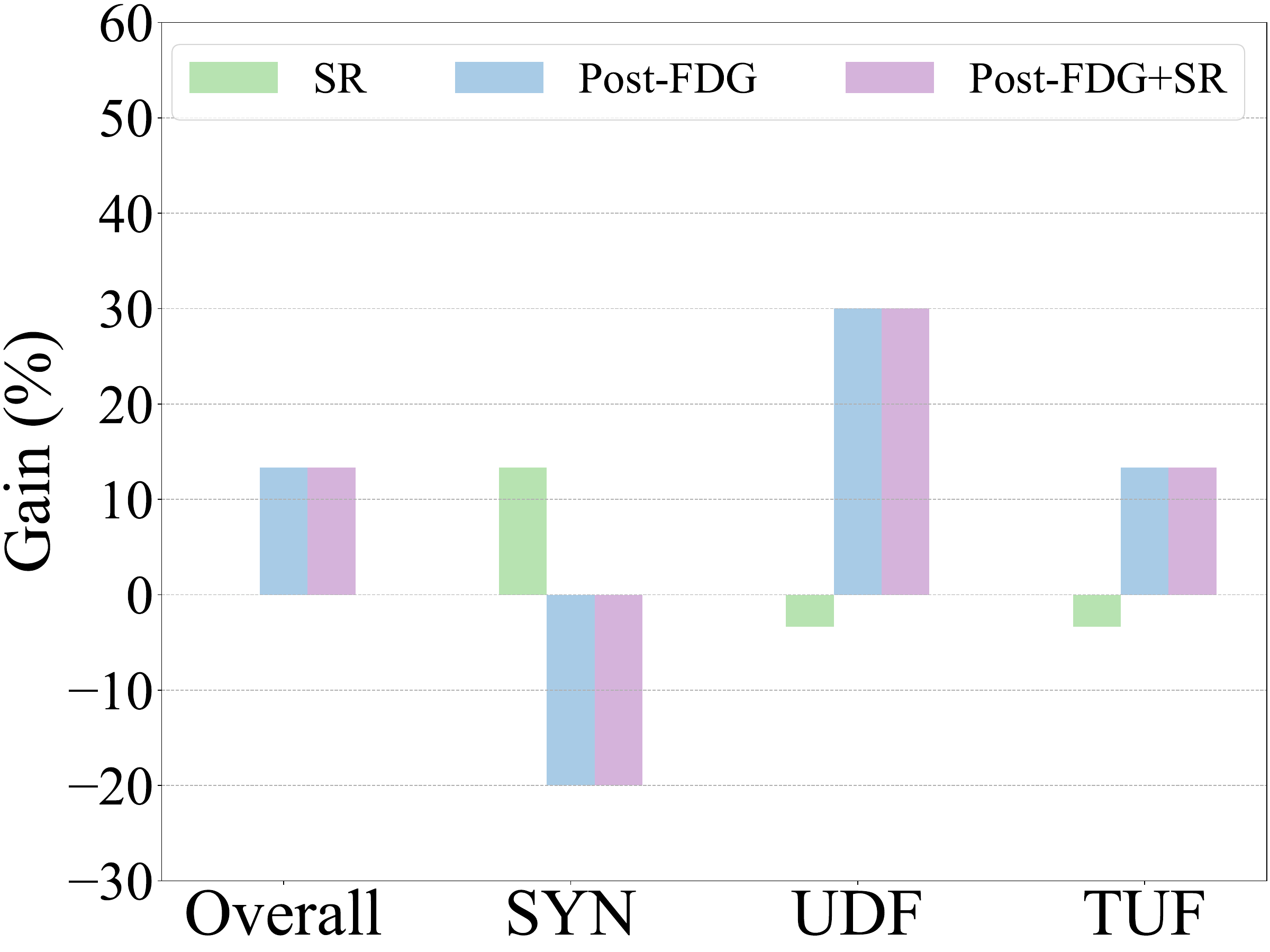}
      \caption{ZS+Detailed on Def\_ArXiv}
      \label{fig:arxiv_zs_det}
    \end{subfigure}
    \caption{Gain of error rates when testing autoformalization with different methods compared to direct test. We evaluate results on zero-shot autoformalized codes and (zero-shot) formal codes with detailed refinement. Testing variants include: (\textbf{SR}): Symbolic Refinement; (\textbf{Post-FDG}): Postprocessing with Formal Definition Grounding.}
    \label{fig:fdg}
\end{figure*}

\subsection{Symbolic Refinement}
Based on reasons 1 and 2 of SYN errors in Table~\ref{tab:failure}, we defined two rules for implementing Symbolic Refinement: (1) if a symbol in the formal code is likely to be an Isabelle symbol (i.e., it starts with ``\textbackslash<'' but misses ``>''), we add ``>'' at its end to ensure that the symbol follows Isabelle's format; (2) for non-existent symbols of mathematical fonts, we replace them with relevant symbols in Isabelle.

The differences in error rates between our methods and direct testing of unmodified autoformalized code are illustrated as bar charts in Figure~\ref{fig:fdg}.

\paragraph{Symbolic Refinement can effectively reduce SYN errors in the generated formal codes on definition datasets.} In Figures~\ref{fig:wiki_zs} and \ref{fig:wiki_zs_det}, both applying symbolic refinement (SR) alone and in combination with Post-FDG lead to a lower SYN error rate on Def\_Wiki-Test. On Def\_ArXiv, Figures~\ref{fig:arxiv_zs} and \ref{fig:arxiv_zs_det} similarly shows that applying SR alone results in a reduction of SYN errors. These results suggest that SR is an effective approach for addressing SYN errors. On miniF2F-Test, however, SR does not impact the error rates because SR is closely tied to specific error patterns in the dataset.

\subsection{Post-FDG}

For implementing FDG, we first extracted external formal definitions of mathematical items and their sources from the Isabelle/HOL library. Then we filtered the extracted definitions to retain only those likely relevant to the autoformalization task on the datasets. Finally, for each individual instance in Def\_Wiki and Def\_ArXiv, we manually determined which formal definitions should be provided as contextual elements. For miniF2F, we simply selected the definitions of real and complex numbers as the relevant definitions. Post-FDG (FDG via \textit{postprocessing}) explicitly augments the preambles generated by LLMs with the sources of relevant formal definitions in formal libraries. 

\paragraph{Autoformalization performance can be underestimated without including contextual information.} In Figure~\ref{fig:fdg}, without modifying the main body of the formalization, replacing the preambles with possible preambles via Post-FDG directly leads to higher overall syntactic correctness. On miniF2F-Test, this setting only considers sources containing formal definitions of real and complex numbers, yet it increases overall syntactic correctness by more than 40\%.

\paragraph{FDG can reduce the occurrence of errors caused by referring to undefined mathematical objects.} In Figure~\ref{fig:fdg}, the UDF error category has the most significant improvement from Post-FDG. Even when LLMs do not include the exact library that contains relevant mathematical items, they tend to use conventional names for the autoformalization task. By importing the appropriate theory files, these previously undefined items can be linked to the formalization, thereby reducing UDF errors.

\paragraph{Errors in autoformalized codes for definition datasets are more likely to be entangled than those in the miniF2F dataset.} In Figure~\ref{fig:miniF2F_zs} and Figure~\ref{fig:miniF2F_zs_det}, Post-FDG leads to positive performance gains across all error categories. However, in Figures~\ref{fig:wiki_zs}, \ref{fig:arxiv_zs}, \ref{fig:wiki_zs_det} and \ref{fig:arxiv_zs_det}, while UDF error rates decrease, error rates in other categories can increase. A similar trend is observed when applying SR, where a reduction in SYN errors can coincide with increases in errors from the other two categories. This phenomenon suggests that because definition datasets are more complex, LLMs are more prone to generating errors from different categories in one code block during the autoformalization process.

\subsection{Generalizability to Lean4}
\begin{table}[!t]
    \small
    \centering
    \begin{tabular}{l | c c c}
        \toprule
        Prompt Strategy & miniF2F & Def\_Wiki & Def\_Arxiv\\
        \midrule
        ZS & 13.93 & 5.36 & 0.00\\
        \midrule
        (ZS)+D & 15.98 & 7.14 & 6.67\\
        \midrule
        (ZS)+D+CR-SYN & 16.80 & 8.93 & 6.67\\
        \midrule
        (ZS)+D+CR-UDF & 13.52 & 10.71 & 6.67\\
        \midrule
        (ZS)+D+CR-TUF & 15.57 & 10.71 & 6.67\\
        \bottomrule
    \end{tabular}
    \caption{GPT-4o pass rates with Lean4. (\textbf{(ZS)+D}): detailed refinement on (zero-shot) formal codes.}
    \label{tab:lean}
\end{table}

We further explore the generalizability of our methods to formal languages beyond Isabelle/HOL. We select Lean4~\citep{lean} as a representative target due to its increasing popularity and widespread use. The Categorical Refinement (CR) method can be applied to Lean4 with only minor prompt modifications. In contrast, exploring the generalizability of Symbolic Refinement and Formal Definition Grounding requires system-specific designs, which fall outside the scope of this paper. We investigate the generalizability of CR on Lean4 and report the pass rates across three datasets using various strategies in Table~\ref{tab:lean}.

\paragraph{Definitions present greater complexity for autoformalization in Lean4.} The pass rates of GPT-4o on definitions are consistently lower than those on miniF2F. For example, on Def\_Arxiv, GPT-4o fails to correctly formalize any of the 30 definitions. Moreover, we observe that providing error details to revise the output improves pass rates, aligning with our observations in Isabelle/HOL.

\paragraph{Categorical Refinement generalizes to Lean4.} On the miniF2F dataset, CR-SYN achieves the highest pass rate. Similarly, on the Def\_Wiki dataset, CR-UDF and CR-TUF yield the best results. These findings demonstrate the effectiveness of the proposed CR in other formal languages. Notably, since the Lean4 assistant does not provide explicit categories as Isabelle/HOL, performance differences across CR categories may indirectly reflect distinct types of errors encountered during autoformalization.

\section{Related Work}

Autoformalization allows for a systematic connection between material and formal inferences~\cite{quan-etal-2024-enhancing,quan-etal-2024-verification}, also enabling the universalization of formal mathematical reasoning. For instance, proof autoformalization has been used as an intermediate step in automated theorem proving~\citep{jiang2023draft,tarrach2024more}. Deep learning models, such as transformers, have been applied to autoformalization in Coq~\citep{cunningham-etal-2022-towards}. In recent years, with the increasing capabilities of LLMs, prompting-based methods have also demonstrated the ability to autoformalize mathematical statements in Isabelle~\citep{wu2022autoformalization,zhang-etal-2024-consistent,li2024autoformalize} and Lean~\citep{yang2023leandojotheoremprovingretrievalaugmented,lu2024processdrivenautoformalizationlean4,liu2025rethinking}. Despite recent progress in autoformalization with LLMs, few studies have analyzed this task from an error perspective. Our work takes a step in this direction.

There are a few benchmarks that provide informal–formal mathematical statement pairs. MiniF2F~\citep{zheng2022miniff} and ProofNet benchmark~\citep{azerbayev2023proofnet} include samples paired with ground-truth formal statements ranging from high-school and undergraduate problems to Olympiad-level problems. However, such informal–formal pairs remain scarce. A growing trend is the development of data generation pipelines for constructing large-scale parallel corpora to finetune LLMs for autoformalization~\citep{jiang2024multilanguage,liu2025atlasautoformalizingtheoremslifting}. While useful, these benchmarks still focus on exercise-style mathematical problems, which do not fully reflect real-world scenarios. In contrast, our definition datasets emphasize real-world statements.

\section{Conclusion}
This paper explored the challenges and advancements in autoformalization of complex mathematical statements. To this end, two datasets collecting real-world definitions in machine learning were introduced for systematic evaluation. By assessing autoformalization performance across three modern LLMs on newly introduced datasets, we identify key failure patterns including syntactic inconsistency, undefined references, and type mismatch. To address these, we proposed interventions such as Categorical Refinement and Formal Definition Grounding to enhance performance. Our results suggest that while modern LLMs exhibit potential in converting natural mathematical definitions into formal representations, they still require improved guidance mechanisms and structured refinement techniques to enhance accuracy. Future research could focus on improving self-correction capabilities and integrating more robust contextual understanding into LLM-based formalization systems.

\section{Limitations}
Despite its contributions, this study has several limitations. First, the error analysis was conducted in Isabelle/HOL, and some results may not directly generalize to other formal proof assistants such as Lean. Second, the definition datasets proposed, though diverse, are relatively small scale. Additionally, while the proposed refinements improve formalization performance, they do not fully eliminate semantic inconsistencies between natural language definitions and their formalized counterparts. More advanced methods are still needed to be developed.

\section*{Acknowledgements}
This work was funded in part by the Swiss National Science Foundation (SNSF/FAPESP) ``RATIONAL'' project.

\bibliography{custom, anthology}

\appendix

\section{Detailed Information about the Dataset Creation}\label{app:data}

We obtain mathematical definitions in the machine learning domain from two sources: Wikipedia (Def\_Wiki) and Arxiv Papers (Def\_ArXiv). For Def\_Wiki, definitions are collected from pages under the Machine Learning category\footnote{\url{https://en.wikipedia.org/wiki/Category:Machine_learning}} and its sub-categories. We manually browsed each page, identified well-defined definitions (i.e., formal descriptions with mathematical symbols), and converted the chosen definitions into LaTeX format. In total, we obtained 56 qualified natural language definitions in LaTeX and divided them into development and test sets, containing 10 and 46 samples, respectively. For Def\_ArXiv, we used the advanced search tool on ArXiv’s website, filtering for papers in the subcategories cs.LG and stat.ML, with comments including "ICML." We restricted the search to papers published in 2019, 2020, and 2021 and manually reviewed the first 25 papers from each year. We shortlisted papers that were accepted to the ICML conference and contained formally described definitions with mathematical symbols to ensure reliability. We then filtered out definitions that were less straightforward or formal in their expressions, extracted the LaTeX versions, and ultimately obtained 30 definitions from 7 papers. We provide dataset statistics in Table~\ref{tab:data_property}.

\begin{table}[!t]
    \scriptsize
    \centering
    \begin{tabular}{c c c c}
        \toprule
        Property & miniF2F-Test & Def\_Wiki & Def\_ArXiv\\
        \midrule
        No. Samples & 244 & 56 & 30\\
        No. Tokens & 70.25 (47.70) & 200.18 (112.98) & 164.40 (71.47)\\
        No. Objects &  4.76 (1.68)& 7.63 (2.71) & 7.10 (2.64)\\
        No. Formulae & 2.71 (1.74) & 2.84 (2.05)& 3.17 (1.97)\\
        \bottomrule
    \end{tabular}
    \caption{Dataset properties. The number of tokens per sample is calculated using the GPT-2 tokenizer. The number of directly mentioned mathematical objects—excluding explicit numbers and variables—and the number of mathematical formulae per sample are estimated through prompting with GPT-4o. The mean (standard deviation) is reported for each dataset.}
    \label{tab:data_property}
\end{table}

\section{Case Study for Formal Definition Grounding}

The following example shows an example of using GPT-4o in a zero-shot setting to formalize the definition of Bradley--Terry model\footnote{\url{https://en.wikipedia.org/wiki/Bradley\%E2\%80\%93Terry_model}}.


\paragraph{Definition of Bradley--Terry model:} Given a pair of items \(i\) and \(j\) drawn from some population, the Bradley--Terry model estimates the probability that the pairwise comparison turns out true, as \[\Pr(i > j) = \frac{p_i}{p_i + p_j}\] where \(p_i\) is a positive real-valued score assigned to individual \(i\).

\vspace{-1.5em}
\begin{lstlisting}[style=isabelle]
theory test
imports Main
begin
definition bradley_terry :: "real \<Rightarrow> real 
  \<Rightarrow> real" where "bradley_terry p_i p_j = 
    p_i / (p_i + p_j)"
end
\end{lstlisting}

The preamble in the generated formal code is ``Main''. However, ``Main'' does not contain the formalization of ``real'', making the formal code invalid. After applying Post-FDG, the preamble is updated to ``HOL.Real'', and the formal code becomes valid. One might suggest creating a universal preamble that imports all source files from the library, applying this common preamble to solve such issues. However, this approach would not align with how a human expert would perform formalization. This failure to identify the correct preambles exposes limitations in the autoformalization capabilities of LLMs. Another issue, which is outside the scope of this paper but an important future direction, is that while Post-FDG can correct the formal code, the semantics of the generated code still do not fully match the original natural language version. For instance, the term ``probability'' does not appear in the formal code, and the phrase ``$p_i$'' is a positive real number” is omitted. 

We acknowledge the importance of quantitatively evaluating semantic consistency in autoformalization. In this paper we mainly quantitatively evaluate the syntactic aspects of formalizations because syntactic correctness can be systematically and fully validated through the theorem prover. However, the full evaluation of semantic consistency still presents technical challenges and remains an open research question. Additionally, it is important to notice that semantic validity is not completely disentangled from syntactic checks. Some important semantic aspects, in fact, are explicitly or implicitly covered within our evaluation. It can be noted, for example, that some of the errors in Table 5 also cover semantic aspects. For instance, a model that is unaware of undefined items will also likely lack a complete semantic understanding of the crucial elements required for correct autoformalization. We believe that coming up with a reliable quantitative method to systematically assess semantic alignment is an important research question to explore in future work, which is out of scope for the current paper.

\section{Examples of Incorrect Formal Codes}

We provide some examples of incorrect formal codes generated by GPT-4o to support our summarized reasons in Table~\ref{tab:failure}. All examples of definitions are from Def\_Wiki development set.

Example 1 is about autoformalization on the definition of Rademacher Complexity\footnote{\url{https://en.wikipedia.org/wiki/Rademacher_complexity}}. In this example, the missing of ``>'' in ``\textbackslash<sigma''leads to a SYN error, which contributes to reason 1 of SYN category. In addition, this piece of formal code uses ``real'' but ``real'' is not imported by ``imports Main''. This problem contributes to reason 1 of UDF category.

\paragraph{Example 1:}

\paragraph{Definition of Rademacher Complexity:} Given a set \(A\subseteq \mathbb{R}^m\), the Rademacher complexity of A is defined as follows: \[\operatorname{Rad}(A):=\frac{1}{m}\mathbb{E}_\sigma\left[\sup_{a \in A}\sum_{i=1}^m \sigma_i a_i\right]\] where \(\sigma_1, \sigma_2, \dots, \sigma_m\) are independent random variables drawn from the Rademacher distribution (i.e. \(\Pr(\sigma_i = +1) = \Pr(\sigma_i = -1) = 1/2\) for \(i=1,2,\dots,m\)), and \(a=(a_1,\dots,a_m)\).

\vspace{-1.5em}
\begin{lstlisting}[style=isabelle]
theory test
imports Main
begin
definition rademacher_distribution :: 
  "nat \<Rightarrow> real pmf" where 
    "rademacher_distribution m = 
     pmf_of_set {x. x = 1 \<or> x = -1}"

definition rademacher_complexity :: 
  "(real vector set) \<Rightarrow> nat \<Rightarrow> real" 
  where "rademacher_complexity A m = 
    (1 / real m) * (expectation 
    (pmf_of_set {x. x = 1 \<or> x = -1}) 
    (\<lambda>\<sigma>. Sup {sum (\<lambda>i. \<sigma i * a i) 
    {1..m} | a. a \<in> A}))"
end
\end{lstlisting}

Example 2 is about autoformalization on the definition of Algorithm Selection\footnote{\url{https://en.wikipedia.org/wiki/Algorithm_selection}}. In this example, the problem of missing ``>'' and using ``real'' also occurs. Even if missing ``>'' is solved, ``\textbackslash<mathcal> is not a valid symbol in Isabelle. This problem contributes to reason 2 of SYN category. In addition, ``instance'' is a reserved keyword in Isabelle so it cannot be used in ``type\_synonym''. This contributes to reason 3 of SYN category.

\paragraph{Example 2:}

\paragraph{Definition of Algorithm Selection:} Given a portfolio \(\mathcal{P}\) of algorithms \(\mathcal{A} \in \mathcal{P}\), a set of instances \(i \in \mathcal{I}\) and a cost metric \(m: \mathcal{P} \times \mathcal{I} \to \mathbb{R}\), the algorithm selection problem consists of finding a mapping \(s: \mathcal{I} \to \mathcal{P}\) from instances \(\mathcal{I}\) to algorithms \(\mathcal{P}\) such that the cost \(\sum_{i \in \mathcal{I}} m(s(i),i)\) across all instances is optimized.

\vspace{-1.5em}
\begin{lstlisting}[style=isabelle]
theory test
imports Main
begin

type_synonym 'a instance = 'a
type_synonym 'a algorithm = 'a
type_synonym cost_metric = "'a algorithm 
  \<Rightarrow> 'a instance \<Rightarrow> real"

definition algorithm_selection ::
  "'a algorithm set \<Rightarrow> 'a instance set 
  \<Rightarrow> cost_metric \<Rightarrow> ('a instance 
  \<Rightarrow> 'a algorithm) \<Rightarrow> real" where
  "algorithm_selection \<mathcal{P} 
    \<mathcal{I} m s = (\<Sum>i \<in> 
    \<mathcal{I}. m (s i) i)"
end
\end{lstlisting}

Example 3 is about autoformalization on the definition of Set-Family Growth Function\footnote{\url{https://en.wikipedia.org/wiki/Growth_function}}. In this example, there is no SYN or UDF errors. The error occurs in the definition of ``growth\_function'', where ``card ()'' is used on the right side of operator ``$\longrightarrow$''. However, the left and right sides of operator ``$\longrightarrow$'' are explicitly defined as type ``bool'' and the type of ``card ()'' is explicitly defined as ``nat'' in the Isabelle/HOL library. This incorrect usage of them causes a type mismatch and leads to TUF error.

\paragraph{Example 3:}

\paragraph{Definition of Set-Family Growth Function:} Let \(H\) be a set family (a set of sets) and \(C\) a set. Their intersection is defined as the following set-family: \(H\cap C := \{h\cap C\mid h\in H\}\). The intersection-size (also called the index) of \(H\) with respect to \(C\) is \(|H\cap C|\). The growth function measures the size of \(H\cap C\) as a function of \(|C|\). Formally: \(\operatorname{Growth}(H,m) := \max_{C: |C|=m} |H\cap C|\).

\vspace{-1.5em}
\begin{lstlisting}[style=isabelle]
theory test
imports Main
begin
definition intersection_set_family :: 
  "'a set set \<Rightarrow> 'a set \<Rightarrow> 'a set set" 
  (infixl "\<inter>\<^sub>H" 70) where
  "H \<inter>\<^sub>H C = {h \<inter> C | h. h \<in> H}"

definition intersection_size :: 
  "'a set set \<Rightarrow> 'a set \<Rightarrow> nat" where
  "intersection_size H C = card (
    H \<inter>\<^sub>H C)"

definition growth_function :: "'a set set \<Rightarrow> 
  nat \<Rightarrow> nat" where
  "growth_function H m = (SUP C. card C = 
    m \<and> finite C \<and> (\<forall>h \<in> H. finite h) 
    \<and> finite H \<longrightarrow> card (H \<inter>\<^sub>H C))"
end
\end{lstlisting}

\section{Formal Definition Grounding via Prompting}

We further explore formal definition grounding with prompting (Prompt-FDG): provide LLMs with grounded formal items and preambles in context to guide autoformalization. We designed two prompts to include external formal definitions for FDG: 1. Soft: allow the LLM some flexibility in whether to use in-context formal definitions for autoformalization; 2. Hard: explicitly instruct the LLM to use the in-context formal definitions if they are related. We tested these prompts on GPT-4o and Def\_Wiki-Test to evaluate whether it can correctly refer to formalised items in context. The results are reported in Table~\ref{tab:wiki_fdc}.

\paragraph{Including relevant formal definitions in the prompt does not boost the performance of autoformalization.} Intuitively, LLMs should perform better when more relevant information is provided within the prompt. However, directly including grounded formal definitions does not positively impact the formalization. This behaviour indicates that current state-of-the-art LLMs cannot effectively link to relevant in-context formal items for autoformalization. How to successfully leverage in-context formal definitions for autoformalization with LLMs remains an important open research question.

\section{Prompts and Additional Results}
The prompts used for the estimation of dataset statistics are provided in Table~\ref{tab:prompt}. The instructions used in the prompts of experiments are provided in Table~\ref{tab:inst_1}. Detailed numbers of autoformalization results on miniF2F test set, Def\_Wiki test set and Def\_ArXiv are provided in Table~\ref{tab:miniF2F_1}, \ref{tab:wiki_1}, \ref{tab:arxiv_1}, respectively. Symbolic refinement results and Post-FDG results on Def\_Wiki test set are provided in Table~\ref{tab:sr} and Table~\ref{tab:wiki_fdc_full}, respectively.

\begin{table}[!t]
    \centering
    \small
    \begin{tabular}{l l | c c c}
        \toprule
        Prompt Strategy & Pass$\uparrow$ & SYN$\downarrow$ & UDF$\downarrow$ & TUF$\downarrow$\\
        \midrule
        ZS & 34.78 & 30.43 & 17.39 & 23.91\\
        \midrule
        Soft-IFDC & 19.57 & 34.78 & 30.43 & 26.09\\
        \midrule
        Hard-IFDC & 19.57 & 36.96 & 21.74 & 39.13\\
        \bottomrule
    \end{tabular}
    \caption{GPT-4o Error results of Prompt-FDG on Def\_Wiki-Test with Post-FDG applied. \textbf{IFDC}: provide LLM with formal definition codes from FDG and force (\textbf{Hard}) or not force (\textbf{Soft}) LLM to use them.}
    \label{tab:wiki_fdc}
\end{table}

\begin{table*}[t]
    \centering
    \small
    \begin{tabular}{p{0.2\textwidth} p{0.72\textwidth}}
        \toprule
        Purpose & Content\\
        \hline
        Mathematical Objects & Given the following statement written in LaTeX: \{\{latex\}\} How many mathematical objects excluding explicit numbers and variables are mentioned directly in this statement? You can think it step by step. Give me the final number as NUMBER=\{the number\}\\
        \hline
        Mathematical Formulae & Given the following statement written in LaTeX: \{\{latex\}\} How many mathematical formulae are mentioned directly in this statement? You can think it step by step. Give me the final number as NUMBER=\{the number\}\\
        \bottomrule
    \end{tabular}
    \caption{Prompts for the estimation of dataset statistics.}
    \label{tab:prompt}
\end{table*}

\begin{table*}[t]
    \centering
    \small
    \begin{tabular}{p{0.12\textwidth} p{0.8\textwidth}}
        \toprule
        Instruction & Content\\
        \hline
        General & You are an expert in \emph{machine learning} and \emph{formal language Isabelle/HOL}. Given the following definition in LaTeX: \{\{latex\}\}, your task is to provide the formal code of this definition in Isabelle/HOL. The following text might contain some preliminaries to explain the given definition: \{\{preliminary\}\}. In case that you need to import any necessary dependent theory files, you should not import any fake theory files.\\
        \hline
        Stylistic & To represent the math symbols, you must use the textual full name of symbols in Isabelle instead of direct symbols. For example you should use \textbackslash<Rightarrow> instead of $\Rightarrow$, \textbackslash<lambda> instead of $\lambda$.\\
        \hline
        Output & Give the results directly without any additional explanations.\\
        \hline
        Refinement & \textbf{Plain}: For your reference, there are some previous formal codes generated by you: \{\{previous\}\}. You can choose to refine this piece of code for your task.\newline
        \textbf{Binary}: For your reference, there are some previous formal codes generated by you: \{\{previous\}\}. The syntactic correctness for this piece of code is: \{\{correctness\}\}. You can choose to refine this piece of code for your task.\newline
        \textbf{Detailed}: For your reference, there are some previous formal codes generated by you: \{\{previous\}\}. The provided code might have some errors according to the Isabelle prover. The error details and where the error code is located in the code are: \{\{error\_details\}\}. You should refine this piece of code for your task.\\
        \hline
        SYN & You should make sure that every symbol you use is a valid Isabelle symbol. If an Isabelle symbol starts with \textbackslash<, then it must end with >. Isabelle reserves some words as keywords. You should be careful with this and avoid to use them to define new names. You should make sure that the usage of symbols and operators is correct in your final output as the incorrect usage will lead to syntax errors.\\
        \hline
        UDF & You should make sure that every item you mentioned in your code has a clear reference either in the local context or the theory files that you decide to import.\\
        \hline
        TUF & You should make sure that in your code, the types of operands of operators or the types of parameters of functions match the types in their definitions exactly. Failure to maintain such compatibility will lead to type mismatch errors.\\
        \hline
        Include Formal Definition Codes & \textbf{Soft}: You can use the following Isabelle/HOL codes to support your task: \{\{formal\_defs\}\} but you should not restate these codes in your final output. You need to formalize everything that is not provided in the given code. In this case, you should assume that you can only use things from HOL.Main. You only need to provide the main body of formal codes for the given definition. You may not import any theory files.\newline
         \textbf{Hard}: The following Isabelle/HOL codes define some mathematical concepts which might be related to your task: \{\{formal\_defs\}\}. If a mathematical concept in your task has been defined in the above codes, you are required to use this version of formal codes but you should not restate these codes in your final output. You need to formalize everything that is not provided in the given code. In this case, you should assume that you can only use things from HOL.Main. You only need to provide the main body of formal codes for the given definition. You may not import any theory files.\\
        \bottomrule
    \end{tabular}
    \caption{Instructions used in prompts.}
    \label{tab:inst_1}
\end{table*}

\begin{table*}[!t]
    \centering
    \small
    \begin{tabular}{l l c c| c c c c c}
        \toprule
        Prompt Strategy & Preamble & Pass$\uparrow$ & FEO$\uparrow$ & TRO$\downarrow$ & IVI$\downarrow$  & SYN$\downarrow$ & UDF$\downarrow$ & TUF$\downarrow$\\
        \midrule
        \multicolumn{9}{l}{\textbf{\textbf{DeepSeekMath-7B}}}\\
        \midrule
        \multirow{2}{*}{ZS} & Direct & 3.28 & 12.79 & 18.44 & 0.00 & 50.00 & 14.34 & 9.43\\
          & Post-FDG & 12.30 & 23.60 & 15.98 & 0.00 & 47.13 & 1.23 & 9.02\\
        \midrule
        \multirow{2}{*}{(ZS) + Binary} & Direct & 2.05 & 6.73 & 2.46 & 0.00 & 79.91 & 5.33 & 2.05\\
          & Post-FDG & 4.10 & 9.39 & 2.46 & 0.00 & 80.33 & 0.41 & 1.23\\
        \midrule
        \multirow{2}{*}{(ZS) + Detailed} & Direct & 3.28 & 10.03 & 5.74 & 0.00 & 70.49 & 10.66 & 4.10\\
          & Post-FDG & 5.74 & 15.57 & 5.74 & 0.00 & 69.67 & 0.82 & 0.41\\
        \midrule
        \multirow{2}{*}{(ZS) + Detailed + CR-All} & Direct & 3.28 & 9.11 & 6.15 & 0.00 & 73.77 & 6.15 & 3.28\\
          & Post-FDG & 5.33 & 13.08 & 6.15 & 0.00 & 72.95 & 0.41 & 3.28\\
        \midrule
        \multicolumn{9}{l}{\textbf{Llama3-8B}}\\
        \midrule
        \multirow{2}{*}{ZS} & Direct & 4.92 & 20.70 & 4.51 & 0.41 & 29.51 & 38.52 & 18.85\\
          & Post-FDG & 10.66 & 31.17 & 4.92 & 0.00 & 28.69 & 20.08 & 21.31\\
        \midrule
        \multirow{2}{*}{(ZS) + Binary} & Direct & 3.69 & 20.52 & 3.28 & 0.41 & 33.20 & 39.75 & 20.49\\
          & Post-FDG & 9.43 & 30.57 & 3.69 & 0.00 & 31.97 & 22.95 & 22.13\\
        \midrule
        \multirow{2}{*}{(ZS) + Detailed} & Direct & 4.10 & 24.33 & 3.69 & 0.82 & 29.51 & 35.25 & 18.44\\
          & Post-FDG & 9.02 & 33.36 & 4.10 & 0.00 & 27.46 & 18.44 & 22.13\\
        \midrule
        \multirow{2}{*}{(ZS) + Detailed + CR-All} & Direct & 4.92 & 24.16 & 6.97 & 0.82 & 27.46 & 35.25 & 20.08\\
          & Post-FDG & 9.43 & 32.41 & 7.79 & 0.00 & 27.46 & 18.85 & 22.54\\
        \midrule
        \multicolumn{9}{l}{\textbf{GPT-4o}}\\
        \midrule
        \multirow{2}{*}{ZS} & Direct & 25.41 & 48.90 & 1.23 & 1.23 & 6.15 & 23.77 & 7.38\\
         & Post-FDG & 67.21 & 81.88 & 0.00 & 0.00 & 3.28 & 2.87 & 5.33\\
        \midrule
        \multirow{2}{*}{ZS + CR-SYN} & Direct & 24.18 & 45.31 & 2.46 & 0.00 & 9.02 & 27.46 & 7.79\\
         & Post-FDG & 52.46 & 73.96 & 0.41 & 0.00 & 7.79 & 3.69 & 3.69\\
        \midrule
        \multirow{2}{*}{ZS + CR-UDF} & Direct & 25.82 & 50.75 & 2.05 & 2.46 & 6.56 & 22.54 & 6.97\\
         & Post-FDG & 61.48 & 80.41 & 0.41 & 0.00 & 5.33 & 2.87 & 2.87\\
        \midrule
        \multirow{2}{*}{ZS + CR-TUF} & Direct & 27.87 & 50.62 & 2.05 & 1.64 & 5.33 & 26.64 & 5.74\\
         & Post-FDG & 54.10 & 78.79 & 0.00 & 0.00 & 3.28 & 4.10 & 2.87\\
        \midrule
        \multirow{2}{*}{(ZS)} & Direct & 25.41 & 53.15 & 1.64 & 1.23 & 6.56 & 22.13 & 7.79\\
         & Post-FDG & 67.21 & 84.05 & 0.00 & 0.00 & 3.28 & 2.46 & 4.92\\
        \midrule
        \multirow{2}{*}{(ZS) + Binary} & Direct & 29.10 & 53.90 & 2.05 & 1.23 & 6.15 & 21.72 & 8.20\\
         & Post-FDG & 67.21 & 83.60 & 0.00 & 0.00 & 4.10 & 2.05 & 4.92\\
        \midrule
        \multirow{2}{*}{(ZS) + Detailed} & Direct & 37.30 & 63.28 & 2.05 & 1.23 & 5.74 & 9.02 & 8.61\\
          & Post-FDG & 83.61 & 91.47 & 0.00 & 0.00 & 2.05 & 0.82 & 3.28\\
        \midrule
        \multirow{2}{*}{(ZS) + CR-SYN} & Direct & 25.41 & 52.72 & 2.05 & 1.23 & 5.74 & 22.13 & 8.61\\
         & Post-FDG & 67.21 & 83.73 & 0.00 & 0.00 & 2.87 & 2.46 & 5.74\\
        \midrule
        \multirow{2}{*}{(ZS) + CR-UDF} & Direct & 26.64 & 54.06 & 1.64 & 1.23 & 6.15 & 21.72 & 6.97\\
         & Post-FDG & 67.21 & 83.78 & 0.00 & 0.00 & 3.69 & 2.05 & 4.92\\
         \midrule
        \multirow{2}{*}{(ZS) + CR-TUF} & Direct & 25.41 & 51.18 & 2.46 & 1.23 & 6.56 & 24.18 & 7.38\\
         & Post-FDG & 67.21 & 83.94 & 0.00 & 0.00 & 3.28 & 2.87 & 4.10\\
        \midrule
        \multirow{2}{*}{(ZS) + Detailed + CR-SYN} & Direct & 38.52 & 64.42 & 2.05 & 1.23 & 7.79 & 8.20 & 7.79\\
          & Post-FDG & 82.79 & 90.32 & 0.00 & 0.00 & 3.28 & 0.82 & 2.05\\
        \midrule
        \multirow{2}{*}{(ZS) + Detailed + CR-UDF} & Direct & 38.11 & 63.95 & 2.05 & 2.46 & 5.74 & 6.56 & 6.97\\
          & Post-FDG & 82.38 & 90.48 & 0.00 & 0.00 & 2.46 & 1.23 & 2.87\\
        \midrule
        \multirow{2}{*}{(ZS) + Detailed + CR-TUF} & Direct & 41.39 & 64.76 & 3.28 & 1.23 & 6.15 & 11.07 & 6.15\\
          & Post-FDG & 83.20 & 90.71 & 0.00 & 0.00 & 2.87 & 1.64 & 2.05\\
        \midrule
        \multirow{2}{*}{(ZS) + Detailed + CR-All} & Direct & 38.52 & 65.73 & 2.05 & 1.23 & 6.15 & 5.74 & 7.79\\
          & Post-FDG & 81.97 & 90.65 & 0.00 & 0.00 & 2.46 & 0.41 & 2.46 \\
        \bottomrule
    \end{tabular}
    \caption{Error results on miniF2F test set.}
    \label{tab:miniF2F_1}
\end{table*}

\begin{table*}[!t]
    \centering
    \small
    \begin{tabular}{l l c c| c c c c c}
        \toprule
        Prompt Strategy & Preamble & Pass$\uparrow$ & FEO$\uparrow$ & TRO$\downarrow$ & IVI$\downarrow$  & SYN$\downarrow$ & UDF$\downarrow$ & TUF$\downarrow$\\
        \midrule
        \multicolumn{9}{l}{\textbf{\textbf{DeepSeekMath-7B}}}\\
        \midrule
        \multirow{2}{*}{ZS} & Direct & 10.87 & 17.75 & 34.78 & 2.17 & 30.43 & 26.09 & 2.17\\
          & Post-FDG & 26.09 & 30.98 & 34.78 & 0.00 & 21.74 & 10.87 & 13.04\\
        \midrule
        \multirow{2}{*}{(ZS) + Binary} & Direct & 6.52 & 7.73 & 8.70 & 0.00 & 69.57 & 21.74 & 2.17\\
          & Post-FDG & 10.87 & 12.56 & 8.70 & 0.00 & 65.22 & 15.22 & 6.52\\
        \midrule
        \multirow{2}{*}{(ZS) + Detailed} & Direct & 10.87 & 13.27 & 15.22 & 2.17 & 43.48 & 34.78 & 6.52\\
          & Post-FDG & 26.09 & 29.21 & 13.04 & 0.00 & 36.96 & 17.39 & 19.57\\
        \midrule
        \multirow{2}{*}{(ZS) + Detailed + CR-All} & Direct & 4.35 & 7.66 & 13.04 & 2.17 & 47.83 & 32.61 & 8.70\\
          & Post-FDG & 17.39 & 21.43 & 13.04 & 0.00 & 41.30 & 15.22 & 21.74\\
        \midrule
        \multicolumn{9}{l}{\textbf{Llama3-8B}}\\
        \midrule
        \multirow{2}{*}{ZS} & Direct & 0.00 & 2.80 & 0.00 & 23.91 & 56.52 & 32.61 & 4.35\\
          & Post-FDG & 0.00 & 2.80 & 21.74 & 0.00 & 58.70 & 23.91 & 15.22\\
        \midrule
        \multirow{2}{*}{(ZS) + Binary} & Direct & 2.17 & 3.71 & 0.00 & 26.09 & 52.17 & 30.43 & 2.17\\
          & Post-FDG & 0.00 & 1.53 & 23.91 & 0.00 & 56.52 & 28.26 & 13.04\\
        \midrule
        \multirow{2}{*}{(ZS) + Detailed} & Direct & 2.17 & 3.80 & 0.00 & 26.09 & 50.00 & 30.43 & 6.52\\
          & Post-FDG & 4.35 & 5.98 & 23.91 & 0.00 & 52.17 & 26.09 & 15.22\\
        \midrule
        \multirow{2}{*}{(ZS) + Detailed + CR-All} & Direct & 2.17 & 3.71 & 0.00 & 26.09 & 52.17 & 32.61 & 4.35\\
          & Post-FDG & 2.17 & 3.71 & 23.91 & 0.00 & 54.35 & 23.91 & 15.22\\
        \midrule
        \multicolumn{9}{l}{\textbf{GPT-4o}}\\
        \midrule
        \multirow{2}{*}{ZS} & Direct & 10.87 & 16.12 & 8.70 & 8.70 & 19.57 & 50.00 & 13.04\\
         & Post-FDG & 34.78 & 42.56 & 6.52 & 0.00 & 30.43 & 17.39 & 23.91\\
        \midrule
        \multirow{2}{*}{ZS + CR-SYN} & Direct & 10.87 & 15.18 & 8.70 & 2.17 & 15.22 & 58.70 & 13.04\\
         & Post-FDG & 34.78 & 40.27 & 8.70 & 0.00 & 28.26 & 13.04 & 26.09\\
        \midrule
        \multirow{2}{*}{ZS + CR-UDF} & Direct & 2.17 & 11.59 & 6.52 & 6.52 & 19.57 & 60.87 & 19.57\\
         & Post-FDG & 30.43 & 42.66 & 2.17 & 0.00 & 34.78 & 23.91 & 23.91\\
        \midrule
        \multirow{2}{*}{ZS + CR-TUF} & Direct & 8.70 & 14.55 & 8.70 & 6.52 & 21.74 & 56.52 & 15.22\\
         & Post-FDG & 30.43 & 40.51 & 6.52 & 0.00 & 34.78 & 17.39 & 28.26\\
        \midrule
        \multirow{2}{*}{(ZS)} & Direct & 10.87 & 16.21 & 8.70 & 8.70 & 19.57 & 50.00 & 13.04\\
         & Post-FDG & 39.13 & 47.23 & 6.52 & 0.00 & 28.26 & 15.22 & 23.91\\
        \midrule
        \multirow{2}{*}{(ZS) + Binary} & Direct & 13.04 & 18.30 & 8.70 & 6.52 & 17.39 & 50.00 & 13.04\\
         & Post-FDG & 39.13 & 48.00 & 6.52 & 0.00 & 26.09 & 8.70 & 28.26\\
        \midrule
        \multirow{2}{*}{(ZS) + Detailed} & Direct & 19.57 & 23.46 & 8.70 & 8.70 & 10.87 & 47.83 & 10.87\\
          & Post-FDG & 43.48 & 50.13 & 6.52 & 0.00 & 21.74 & 10.87 & 23.91\\
        \midrule
        \multirow{2}{*}{(ZS) + CR-SYN} & Direct & 10.87 & 16.12 & 8.70 & 8.70 & 17.39 & 52.17 & 13.04\\
         & Post-FDG & 36.96 & 44.97 & 6.52 & 0.00 & 30.43 & 15.22 & 23.91\\
        \midrule
        \multirow{2}{*}{(ZS) + CR-UDF} & Direct & 10.87 & 16.12 & 8.70 & 8.70 & 19.57 & 50.00 & 13.04\\
         & Post-FDG & 36.96 & 44.97 & 6.52 & 0.00 & 30.43 & 15.22 & 23.91\\
         \midrule
        \multirow{2}{*}{(ZS) + CR-TUF} & Direct & 10.87 & 16.21 & 8.70 & 8.70 & 21.74 & 47.83 & 13.04\\
         & Post-FDG & 36.96 & 45.06 & 6.52 & 0.00 & 32.61 & 15.22 & 21.74\\
        \midrule
        \multirow{2}{*}{(ZS + Detailed) + Detailed} & Direct & 19.57 & 24.09 & 8.70 & 8.70 & 13.04 & 43.48 & 10.87\\
          & Post-FDG & 43.48 & 50.32 & 6.52 & 0.00 & 19.57 & 8.70 & 26.09\\
        \midrule
        \multirow{2}{*}{(ZS) + Detailed + CR-SYN} & Direct & 21.74 & 25.63 & 8.70 & 10.87 & 10.87 & 41.30 & 8.70\\
          & Post-FDG & 45.65 & 52.31 & 6.52 & 0.0 & 21.74 & 8.70 & 21.74\\
        \midrule
        \multirow{2}{*}{(ZS) + Detailed + CR-UDF} & Direct & 17.39 & 21.83 & 8.70 & 13.04 & 17.39 & 39.13 & 8.70\\
          & Post-FDG & 43.48 & 50.24 & 6.52 & 0.0 & 21.74 & 10.87 & 21.74\\
        \midrule
        \multirow{2}{*}{(ZS) + Detailed + CR-TUF} & Direct & 19.57 & 23.46 & 8.70 & 8.70 & 17.39 & 43.48 & 8.70\\
          & Post-FDG & 45.65 & 52.31 & 6.52 & 0.0 & 23.91 & 10.87 & 19.57\\
        \midrule
        \multirow{2}{*}{(ZS) + Detailed + CR-All} & Direct & 21.74 & 25.63 & 8.70 & 8.70 & 10.87 & 43.48 & 13.04\\
          & Post-FDG & 43.48 & 50.13 & 6.52 & 0.00 & 21.74 & 10.87 & 23.91\\
        \bottomrule
    \end{tabular}
    \caption{Error results on Def\_Wiki test set.}
    \label{tab:wiki_1}
\end{table*}

\begin{table*}[!t]
    \centering
    \small
    \begin{tabular}{l l c c| c c c c c c}
        \toprule
        Prompt Strategy & Preamble & Pass$\uparrow$ & FEO$\uparrow$ & TRO$\downarrow$ & IVI$\downarrow$  & SYN$\downarrow$ & UDF$\downarrow$ & TUF$\downarrow$\\
        \midrule
        \multicolumn{9}{l}{\textbf{\textbf{DeepSeekMath-7B}}}\\
        \midrule
        \multirow{2}{*}{ZS} & Direct & 13.33 & 14.69 & 16.67 & 0.00 & 40.00 & 36.67 & 13.33\\
          & Post-FDG & 16.67 & 18.02 & 13.33 & 0.00 & 43.33 & 30.00 & 16.67\\
        \midrule
        \multirow{2}{*}{(ZS) + Binary} & Direct & 3.33 & 3.33 & 6.67 & 0.00 & 66.67 & 33.33 & 3.33\\
          & Post-FDG & 6.67 & 7.41 & 3.33 & 0.00 & 70.00 & 23.33 & 10.00\\
        \midrule
        \multirow{2}{*}{(ZS) + Detailed} & Direct & 6.67 & 7.36 & 13.33 & 0.00 & 46.67 & 43.33 & 13.33\\
          & Post-FDG & 13.33 & 14.02 & 10.00 & 0.00 & 46.67 & 33.33 & 20.00\\
        \midrule
        \multirow{2}{*}{(ZS) + Detailed + CR-All} & Direct & 6.67 & 7.59 & 13.33 & 0.00 & 46.67 & 43.33 & 13.33\\
          & Post-FDG & 13.33 & 14.26 & 10.00 & 0.00 & 46.67 & 33.33 & 20.00\\
        \midrule
        \multicolumn{9}{l}{\textbf{Llama3-8B}}\\
        \midrule
        \multirow{2}{*}{ZS} & Direct & 0.00 & 2.67 & 0.00 & 13.33 & 70.00 & 40.00 & 6.67\\
          & Post-FDG & 0.00 & 2.67 & 13.33 & 0.00 & 66.67 & 26.67 & 20.00\\
        \midrule
        \multirow{2}{*}{(ZS) + Binary} & Direct & 3.33 & 5.83 & 0.00 & 20.00 & 60.00 & 33.33 & 6.67\\
          & Post-FDG & 3.33 & 5.83 & 20.00 & 0.00 & 60.00 & 26.67 & 16.67\\
        \midrule
        \multirow{2}{*}{(ZS) + Detailed} & Direct & 0.00 & 1.41 & 0.00 & 20.00 & 63.33 & 33.33 & 6.67\\
          & Post-FDG & 0.00 & 4.22 & 20.00 & 0.00 & 56.67 & 26.67 & 20.00\\
        \midrule
        \multirow{2}{*}{(ZS) + Detailed + CR-All} & Direct & 0.00 & 2.33 & 0.00 & 16.67 & 66.67 & 36.67 & 6.67\\
          & Post-FDG & 3.33 & 7.00 & 16.67 & 0.00 & 63.33 & 26.67 & 23.33\\
        \midrule
        \multicolumn{9}{l}{\textbf{GPT-4o}}\\
        \midrule
        \multirow{2}{*}{ZS} & Direct & 13.33 & 19.30 & 0.00 & 0.00 & 40.00 & 56.66 & 6.67\\
         & Post-FDG & 23.33 & 36.02 & 0.00 & 0.00 & 60.00 & 13.33 & 13.33\\
        \midrule
        \multirow{2}{*}{ZS + CR-SYN} & Direct & 10.00 & 17.14 & 0.00 & 0.00 & 26.67 & 66.67 & 6.67\\
         & Post-FDG & 26.67 & 39.11 & 0.00 & 0.00 & 50.00 & 20.00 & 16.67\\
        \midrule
        \multirow{2}{*}{ZS + CR-UDF} & Direct & 10.00 & 18.54 & 0.00 & 10.00 & 33.33 & 46.67 & 16.67\\
         & Post-FDG & 23.33 & 36.52 & 0.00 & 0.00 & 46.67 & 23.33 & 16.67\\
        \midrule
        \multirow{2}{*}{ZS + CR-TUF} & Direct & 6.67 & 14.05 & 0.00 & 3.33 & 23.33 & 63.33 & 10.00\\
         & Post-FDG & 23.33 & 35.03 & 0.00 & 0.00 & 56.67 & 13.33 & 10.00\\
        \midrule
        \multirow{2}{*}{(ZS)} & Direct & 16.67 & 23.28 & 0.00 & 0.00 & 36.67 & 53.33 & 6.67\\
         & Post-FDG & 30.00 & 40.83 & 0.00 & 0.00 & 56.67 & 10.00 & 10.00\\
        \midrule
        \multirow{2}{*}{(ZS) + Binary} & Direct & 16.67 & 24.30 & 0.00 & 0.00 & 33.33 & 53.33 & 6.67\\
         & Post-FDG & 26.67 & 41.02 & 0.00 & 0.00 & 60.00 & 10.00 & 6.67\\
        \midrule
        \multirow{2}{*}{(ZS) + Detailed} & Direct & 16.67 & 28.91 & 0.00 & 0.00 & 36.67 & 43.33 & 16.67\\
          & Post-FDG & 30.00 & 44.15 & 0.00 & 0.00 & 56.67 & 13.33 & 3.33\\
        \midrule
        \multirow{2}{*}{(ZS) + CR-SYN} & Direct & 20.00 & 24.12 & 0.00 & 0.00 & 36.67 & 53.33 & 3.33\\
         & Post-FDG & 30.00 & 40.83 & 0.00 & 0.00 & 60.00 & 10.00 & 6.67\\
        \midrule
        \multirow{2}{*}{(ZS) + CR-UDF} & Direct & 20.00 & 24.12 & 0.00 & 0.00 & 30.00 & 56.67 & 6.67\\
         & Post-FDG & 30.00 & 40.83 & 0.00 & 0.00 & 56.67 & 10.00 & 10.00\\
         \midrule
        \multirow{2}{*}{(ZS) + CR-TUF} & Direct & 16.67 & 23.07 & 0.00 & 0.00 & 33.33 & 53.33 & 10.00\\
         & Post-FDG & 26.67 & 37.47 & 0.00 & 0.00 & 60.00 & 13.33 & 6.67\\
        \midrule
        \multirow{2}{*}{(ZS) + Detailed + CR-SYN} & Direct & 23.33 & 29.74 & 0.00 & 0.00 & 30.00 & 50.00 & 10.00\\
          & Post-FDG & 30.00 & 43.12 & 0.00 & 0.00 & 53.33 & 16.67 & 3.33\\
        \midrule
        \multirow{2}{*}{(ZS) + Detailed + CR-UDF} & Direct & 26.67 & 34.18 & 0.00 & 0.00 & 33.33 & 43.33 & 10.00\\
          & Post-FDG & 30.00 & 44.23 & 0.00 & 0.00 & 53.33 & 13.33 & 6.67\\
        \midrule
        \multirow{2}{*}{(ZS) + Detailed + CR-TUF} & Direct & 13.33 & 25.41 & 0.00 & 0.00 & 33.33 & 46.67 & 16.67\\
          & Post-FDG & 30.00 & 43.98 & 0.00 & 0.00 & 56.67 & 13.33 & 3.33\\
        \midrule
        \multirow{2}{*}{(ZS) + Detailed + CR-All} & Direct & 13.33 & 24.54 & 0.00 & 0.00 & 33.33 & 50.00 & 16.67\\
          & Post-FDG & 33.33 & 46.45 & 0.00 & 0.00 & 50.00 & 13.33 & 6.67\\
        \bottomrule
    \end{tabular}
    \caption{Error results on Def\_ArXiv set.}
    \label{tab:arxiv_1}
\end{table*}

\begin{table*}[!t]
    \centering
    \small
    \begin{tabular}{l l c c| c c c c c}
        \toprule
        Prompt Strategy & Preamble & Pass$\uparrow$ & FEO$\uparrow$ & TRO$\downarrow$ & IVI$\downarrow$  & SYN$\downarrow$ & UDF$\downarrow$ & TUF$\downarrow$\\
        \midrule
        \multicolumn{9}{l}{\textbf{\textbf{miniF2F-Test}}}\\
        \midrule
        \multirow{2}{*}{ZS} & Direct & 25.41 & 48.90 & 1.23 & 1.23 & 6.15 & 23.77 & 7.38\\
          & Post-FDG & 67.21 & 81.88 & 0.00 & 0.00 & 3.28 & 2.87 & 5.33\\
        \midrule
        \multirow{2}{*}{(ZS) + Detailed} & Direct & 37.30 & 63.28 & 2.05 & 1.23 & 5.74 & 9.02 & 8.61\\
          & Post-FDG & 83.61 & 91.47 & 0.00 & 0.00 & 2.05 & 0.82 & 3.28\\
        \midrule
        \multicolumn{9}{l}{\textbf{Def\_Wiki-Test}}\\
        \midrule
        \multirow{2}{*}{ZS} & Direct & 10.87 & 16.43 & 8.70 & 8.70 & 15.22 & 52.17 & 13.04\\
          & Post-FDG & 34.78 & 43.19 & 6.52 & 0.00 & 23.91 & 19.57 & 28.26\\
        \midrule
        \multirow{2}{*}{(ZS) + Detailed} & Direct & 19.57 & 23.77 & 8.70 & 8.70 & 8.70 & 47.83 & 10.87\\
          & Post-FDG & 43.48 & 50.76 & 6.52 & 0.00 & 17.39 & 10.87 & 28.26\\
        \midrule
        \multicolumn{9}{l}{\textbf{Def\_ArXiv}}\\
        \midrule
        \multirow{2}{*}{ZS} & Direct & 13.33 & 19.30 & 0.00 & 0.00 & 23.33 & 66.67 & 6.67\\
         & Post-FDG & 23.33 & 36.02 & 0.00 & 0.00 & 60.00 & 13.33 & 13.33\\
        \midrule
        \multirow{2}{*}{(ZS) + Detailed} & Direct & 16.67 & 28.91 & 0.00 & 0.00 & 23.33 & 46.67 & 20.00\\
          & Post-FDG & 30.00 & 44.15 & 0.00 & 0.00 & 56.67 & 13.33 & 3.33\\
        \bottomrule
    \end{tabular}
    \caption{Symbolic refinement of GPT-4o results on three dataset.}
    \label{tab:sr}
\end{table*}

\begin{table*}[!t]
    \centering
    \small
    \begin{tabular}{l l c c| c c c c c}
        \toprule
        Prompt Strategy & Preamble & Pass$\uparrow$ & FEO$\uparrow$ & TRO$\downarrow$ & IVI$\downarrow$  & SYN$\downarrow$ & UDF$\downarrow$ & TUF$\downarrow$\\
        \midrule
        \multicolumn{9}{l}{\textbf{GPT-4o}}\\
        \midrule
        \multirow{2}{*}{Soft-IFDC} & Direct & 6.52 & 11.45 & 8.70 & 0.00 & 17.39 & 71.74 & 2.17\\
         & Post-FDG & 19.57 & 29.65 & 0.00 & 0.00 & 34.78 & 30.43 & 26.09\\
        \midrule
        \multirow{2}{*}{Hard-IFDC} & Direct & 4.35 & 11.86 & 10.87 & 0.00 & 10.87 & 69.57 & 6.52\\
          & Post-FDG & 19.57 & 26.95 & 0.00 & 0.00 & 36.96 & 21.74 & 39.13\\
        \midrule
        \multirow{2}{*}{(ZS) + Soft-IFDC + Binary} & Direct & 15.22 & 20.47 & 8.70 & 2.17 & 15.22 & 58.70 & 10.87\\
          & Post-FDG & 41.30 & 51.09 & 6.52 & 0.00 & 26.09 & 10.87 & 26.09\\
        \midrule
        \multirow{2}{*}{(ZS) + Soft-IFDC + Detailed} & Direct & 15.22 & 20.20 & 8.70 & 2.17 & 13.04 & 56.52 & 13.04\\
          & Post-FDG & 41.30 & 51.26 & 6.52 & 0.0 & 23.91 & 10.87 & 26.09\\
        \bottomrule
    \end{tabular}
    \caption{Prompt-FDG results on Def\_Wiki test set.}
    \label{tab:wiki_fdc_full}
\end{table*}

\end{document}